\documentclass[acmsmall,screen,nonacm]{acmart}
\usepackage{mathpartir}
\usepackage{mathtools}
\usepackage[utf8]{inputenc}
\usepackage{dsfont}
\usepackage{fontawesome5}
\usepackage{amsmath,amsthm}
\usepackage{enumitem}
\usepackage{lipsum} %
\usepackage[dvipsnames, prologue]{xcolor}
\usepackage{changebar}
\usepackage{listings} %
\usepackage{fvextra} %
\usepackage{cleveref}

\usepackage{color}
\usepackage{threeparttable}
\usepackage{caption} %
\usepackage{subcaption} %
\usepackage{semantic}
\usepackage{hyperref}
\usepackage{algorithm}
\usepackage{algpseudocode}

\let\AlgoWhile\While
\let\AlgoEndWhile\EndWhile

\let\AlgoState\State
\let\AlgoReturn\Return
\let\AlgoFunction\Function
\let\AlgoEndFunction\EndFunction
\usepackage{framed} %
\usepackage{stmaryrd} %
\usepackage{tikz}
\usepackage{tikz-cd}
\usepackage{array} %
\usetikzlibrary{arrows.meta, shapes.multipart, positioning}
\usepackage{multicol} %
\usepackage{placeins} %
\usepackage{soul} %
\usepackage{pifont} %
\usepackage{wrapfig} %
\usepackage[scaled=0.85]{beramono} %
\usepackage[export]{adjustbox} %
\PassOptionsToPackage{varqu,varl}{zi4}

\definecolor{myMainBlue}{HTML}{56C1FF}
\colorlet{myBlue}{myMainBlue!40!white}
\definecolor{myYellow}{HTML}{FFFC66}
\definecolor{myMainRed}{HTML}{FF968D}
\definecolor{myArrowRed}{HTML}{A31919}
\colorlet{myRed}{myMainRed!40!white}
\definecolor{myGreen}{HTML}{E8FEE2}
\definecolor{myGray}{HTML}{888888}

\definecolor{impComment}{HTML}{1A7A3C} %
\lstdefinelanguage{Imprompt}{
    morekeywords={let,reason,as,if,else,while,get,from,say,skip,true,false},
    keywordstyle=\bfseries\color{BlueViolet},
    morecomment=[l]{//},
    commentstyle=\itshape\bfseries\color{impComment},
    morestring=[b]",
    stringstyle=\itshape\color{myGray},
    sensitive=true,
    showstringspaces=false,
    keepspaces=true,
    columns=fullflexible,
    basicstyle=\ttfamily\scriptsize,
    numberstyle=\ttfamily\scriptsize\color{gray},
    breaklines=true,
}

\newcommand{\C}{\mathcal{C}}
\newcommand{\D}{\mathcal{D}}

\renewcommand{\t}{\boldsymbol{t}}
\newcommand{\Sem}[1]{\llbracket{#1}\rrbracket}
\newcommand{\Tok}{\texttt{Tok}}
\newcommand{\Detok}[1]{\mathsf{Detok}(#1)}
\newcommand{\Decode}[2]{\mathsf{decode}(#1,#2)}
\newcommand{\Comp}{\textsf{Comp}}
\newcommand{\BComp}{\textsf{BComp}}
\newcommand{\SComp}{\textsf{SComp}}
\newcommand{\DSL}{\textsc{Imprompt}}
\newcommand{\DSLIR}{\textsc{Imprompt-IR}}
\newcommand{\DSLPy}{\textsc{Imprompt-Py}}
\newcommand{\IMP}{\textsc{Imp}}
\newcommand{\Paren}[1]{\mathopen{}\left( {#1}_{{}_{}}\,\negthickspace\right)\mathclose{}}
\newcommand{\Brack}[1]{\mathopen{}\left[ {#1}_{{}_{}}\,\negthickspace\right]\mathclose{}}
\newcommand{\Rule}[2]{\hyperref[#1]{\textsc{#2}}}
\newcommand{\Line}[1]{{\color{gray}{\texttt{#1}}}}
\newcommand{\One}{\raisebox{-.18ex}{\ding{182}}}
\newcommand{\Two}{\raisebox{-.18ex}{\ding{183}}}
\newcommand{\Three}{\raisebox{-.18ex}{\ding{184}}}
\newcommand{\Four}{\raisebox{-.18ex}{\ding{185}}}
\newcommand{\Five}{\raisebox{-.18ex}{\ding{186}}}

\newcommand{\Ceil}[1]{\lceil{#1}\rceil}
\newcommand{\Floor}[1]{\lfloor{#1}\rfloor}
\newcommand{\Leadsto}{\rightsquigarrow}
\newcommand{\Typeof}[1]{\textsf{typeof}(#1)}
\newcommand{\MV}[1]{\mathsf{mv}(#1)}

\newcommand{\Code}[1]{{\color{BlueViolet}{\textup{\texttt{\textbf{#1}}}}}}
\newcommand{\IRCode}[1]{{\texttt{\textbf{#1}}}}
\newcommand{\Text}[1]{{\color{myGray}{{\small\textit{\texttt{#1}}}}}}

\newcommand{\Var}{\textsf{Var}}
\newcommand{\Stmt}{\textsf{Stmt}}
\newcommand{\IR}{\textsf{IR}}
\newcommand{\Tmpl}{\textsf{Tmpl}}
\newcommand{\Lit}{\textsf{Lit}}
\newcommand{\Expr}{\textsf{Expr}}

\newcommand{\Store}{\textsf{Store}}
\newcommand{\Int}{\textsf{Int}}
\newcommand{\Type}{\textsf{Type}}

\newcommand{\Bool}{\texttt{Bool}}
\newcommand{\Dyn}{\texttt{Dyn}}

\newcommand{\Ty}[1]{\texttt{Ty<}#1\texttt{>}}

\newcommand{\True}{{\Code{true}}}
\newcommand{\False}{{\Code{false}}}
\newcommand{\As}{{\Code{as}}}
\newcommand{\Colon}{{\Code{:}~}}

\newcommand{\Let}{{\Code{let}}}
\newcommand{\Get}{{\Code{get}}}
\newcommand{\Eq}{\Code{=}}
\newcommand{\Say}{{\Code{say}}}
\newcommand{\Reason}{{\Code{reason}}}
\renewcommand{\If}{{\Code{if}}}
\newcommand{\From}{{\Code{from}}}
\renewcommand{\While}{{\Code{while}}}
\newcommand{\LBrace}{{\Code{\{}}}
\newcommand{\RBrace}{{\Code{\}}}}

\renewcommand{\Else}{{\Code{else}}}
\renewcommand{\And}{{\Code{\&\&}}}
\newcommand{\Or}{{\Code{|\!|}}}
\newcommand{\Not}{{\Code{!}}}
\newcommand{\Skip}{{\Code{skip}}}

\newcommand{\Brace}[1]{\LBrace{\color{black}\textup{#1}}\RBrace}
\newcommand{\LetEqStmt}[2]{\Let~{#1}~\Eq~{#2}}
\newcommand{\SayStmt}[1]{\Say~{#1}}
\newcommand{\ReasonExpr}[1]{\Reason~{#1}}
\newcommand{\CastExpr}[2]{{#1}~\As~{#2}}
\newcommand{\SeqStmt}[2]{{#1}~{#2}}
\newcommand{\IfStmt}[2]{\If~{#1}~\LBrace~{#2}~\RBrace}
\newcommand{\WhileStmt}[2]{\While~{#1}~\LBrace~{#2}~\RBrace}

\newcommand{\ElseStmt}[1]{\Else~\LBrace~{#1}~\RBrace}
\newcommand{\GetStmt}[2]{\Get~{#1}~\From~{#2}}
\newcommand{\Angle}[1]{\left\langle{#1}\right\rangle}

\newcommand{\Push}[1]{\IRCode{push}~{#1}}
\newcommand{\Post}{\IRCode{post}} %
\newcommand{\Gen}{\IRCode{gen}}
\newcommand{\Pop}[1]{\IRCode{pop}~{#1}}
\newcommand{\Cgen}[1]{\IRCode{cgen}~{#1}}
\newcommand{\Print}{\IRCode{print}}
\newcommand{\Cat}{\IRCode{cat}}
\newcommand{\Lookup}[1]{\IRCode{lookup}~{#1}}
\newcommand{\Dup}{\IRCode{dup}}
\newcommand{\TemplateIR}[1]{\IRCode{template}~{#1}}
\newcommand{\Seq}[2]{{#1}\IRCode{;}~{#2}}
\newcommand{\Nop}{\IRCode{nop}}
\newcommand{\Branch}[2]{\IRCode{br}~{#1}~{#2}}
\newcommand{\IRWhile}[2]{\IRCode{while}~{#1}~{#2}}
\newcommand{\AndOp}{\IRCode{and}}
\newcommand{\OrOp}{\IRCode{or}}
\newcommand{\NotOp}{\IRCode{not}}

\newcommand{\Dom}[1]{\operatorname{dom}(#1)}
\newcommand{\Mod}{\operatorname{Mod}}
\newcommand{\llbrace}{\{\kern-0.53ex|}
\newcommand{\rrbrace}{|\kern-0.53ex\}}
\newcommand{\Token}[1]{{\small\texttt{\lbrack\!{#1}}\!\texttt{\rbrack}}}
\newcommand{\EOS}{\Token{\texttt{eos}}}

\begin{document}
\title{$\DSL$: A Language Framework for Prompt Programming}
\author{Chentian Wu}
\email{chentian.wu@wisc.edu}
\author{Shengyuan Yang}
\email{syang686@wisc.edu}
\author{Adithya Murali}
\email{adithyamurali@cs.wisc.edu}
\settopmatter{printacmref=false}
\setcopyright{none}
\begin{abstract}
With the unprecedented success of Language Models (LMs), the science of Prompt Engineering has evolved the powerful idea of \emph{Prompt Programming}, where prompts are treated as a programmable control surface for describing complex tasks and leveraging LM capabilities. However, existing prompt programming frameworks suffer from various complexities and inelegances, which make them hard to utilize in practice for effectively describing tasks. We propose $\DSL$, a new language framework for the study and practice of prompt programming. We undertake a foundational investigation of prompt programming, and contend that prompt programs must contain only the task descriptions and must be decoupled from lower-level `execution' details. We further develop this position by illustrating structured prompting as a combination of prompt programming and prompt program `compilation'. We exemplify this view by formally defining two compilers for $\DSL$ programs. We then explore the idea of typing for prompt programs and draw a correspondence between type checking and constrained decoding. Finally, we implement our compilers and type checkers and evaluate them on a variety of case studies. We believe our work contributes programming-language foundations toward the emerging area of prompt programming.

\end{abstract}

\maketitle
\section{Introduction}\label{sec:intro}

Large Language Models (LLMs) have demonstrated substantial utility across a wide range of language-based tasks~\cite{vaswani2017attention,devlin-etal-2019-bert,Radford2019LanguageMA}.
At a high level, an LM defines a probability distribution over token sequences.
A generative LM takes tokenized input, called a \emph{prompt}, and predicts subsequent tokens conditioned on the prompt and previously generated context.
Prior studies~\cite{brown2020language,gao-etal-2021-making,wei2022chain} have shown that even when users intend to specify the same task,
small prompt variations can lead to large differences in response quality and correctness.
This has motivated extensive work~\cite{zhang2023automatic,wang2023selfconsistencyimproveschainthought,10.5555/3495724.3496517,10.1145/3560815,yao2023reactsynergizingreasoningacting,10.5555/3666122.3666639,sahoo2025systematicsurveypromptengineering,schulhoff2024promptreportsystematicsurvey} on how to design better prompts, commonly referred to as \emph{prompt engineering}.

\paragraph{Prompt Programming}
To make prompt engineering more consistent and reproducible,
many programming frameworks or languages~\cite{langchain,langgraph2025,dspykhattab2024,sglang,guidance2025,
outlines2025,outlineswillard2023efficient,genaiscript2025,promptflow2025,promptpexsharma2025automatictestgeneration,sammo.schnabel-neville-2024-symbolic,lmql10.1145/3591300,llamppllew2023sequentialmontecarlosteering,appldong2024,pdlvaziri2024pdldeclarativepromptprogramming,10.1145/3763143} have been proposed.
These systems help users reuse effective prompt patterns to produce higher-quality prompts, while also providing higher-level syntactic and semantic control over LM outputs for more reliable behavior.
For example, DSPy~\cite{dspykhattab2024} decomposes LM-system development into programming, evaluation, and optimization, supporting modular composition and partially automated prompt optimization.
LMQL~\cite{lmql10.1145/3591300} provides a high-level language with declarative SQL-like elements and imperative scripting constructs, allowing users to specify decoding constraints precisely.

\paragraph{Key Challenge in Prompt Programming}
We identify two core challenges across existing prompt frameworks and languages.

\begin{figure}[htbp!]
    \leftskip=0pt
    \includegraphics[width=0.8\textwidth]{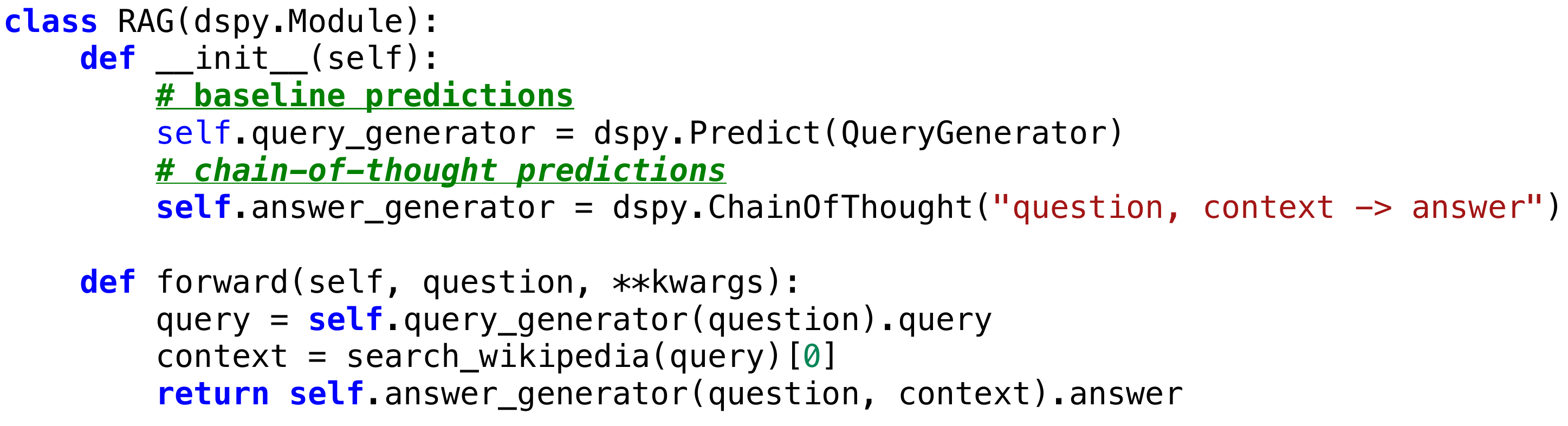}   
    \vspace{1ex}
    \caption{RAG application from DSPy homepage~\cite{dspyhomepage}}
    \label{fig:rag-dspy}
    \vspace{-3ex}
\end{figure}

First, most prompt programming frameworks expose highly detailed prompt APIs, including interfaces that encapsulate prompting techniques such as chain-of-thought~\cite{wei2022chain} and ReAct~\cite{yao2023reactsynergizingreasoningacting}.
This couples task-level prompt description with lower-level `execution' details in the user's code. %
\Cref{fig:rag-dspy} shows a simple Retrieval-Augmented Generation (RAG) application in DSPy.
Users must manually choose between \texttt{dspy.Predict} and \texttt{dspy.ChainOfThought} for each generator.
Such choices are often empirical and can significantly affect performance. In this paper we argue that prompt description must be decoupled from its execution details.
Users should focus on prompt description, 
while the compiler should handle execution details.

Second, most existing prompt languages lack well-defined compilation semantics.
Unlike classical programming languages, 
prompt languages do not have a fixed low-level instruction set.
We therefore cannot assign formal semantics to prompt languages in a direct, meaningful way.
Instead, the semantics of a prompt language is defined by its compilation.
Different compilation strategies induce potentially different semantics, and these semantic differences can directly affect LM behavior.
To the best of our knowledge, 
no prior work on prompt programming has systematically analyzed performance and behavioral differences induced by varying compilation semantics.

\begin{figure}[H]
    \vspace{-0.5ex}
    \centering
    \begin{subfigure}{.48\textwidth}
      \centering
      \includegraphics[width=\linewidth]{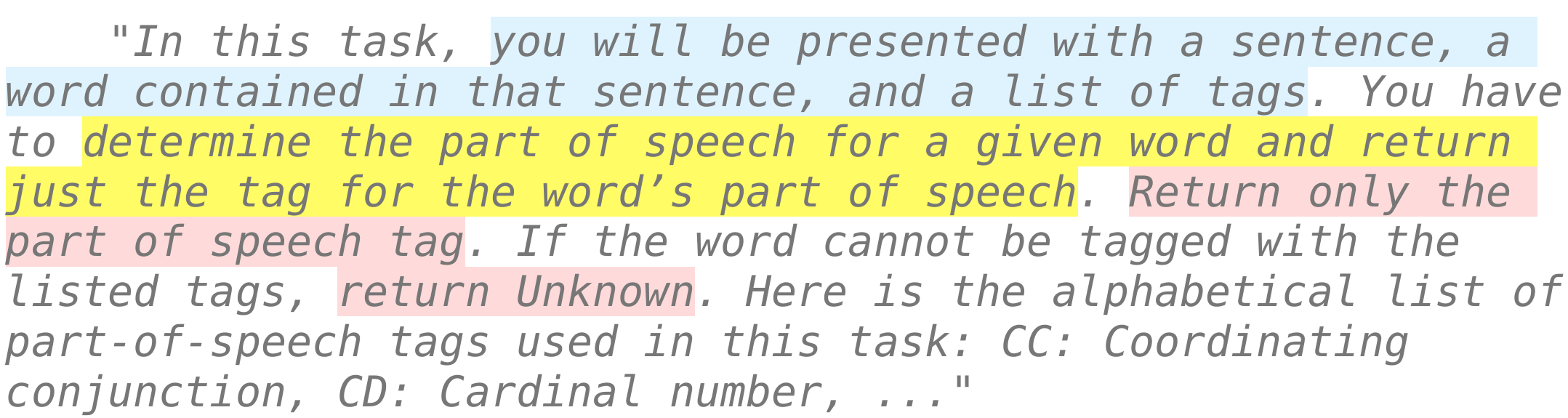}
      \caption{PoS in Natural Language}\label{fig:pos:1}
    \end{subfigure}
    \begin{subfigure}{.50\textwidth}
      \centering
      \includegraphics[width=\linewidth]{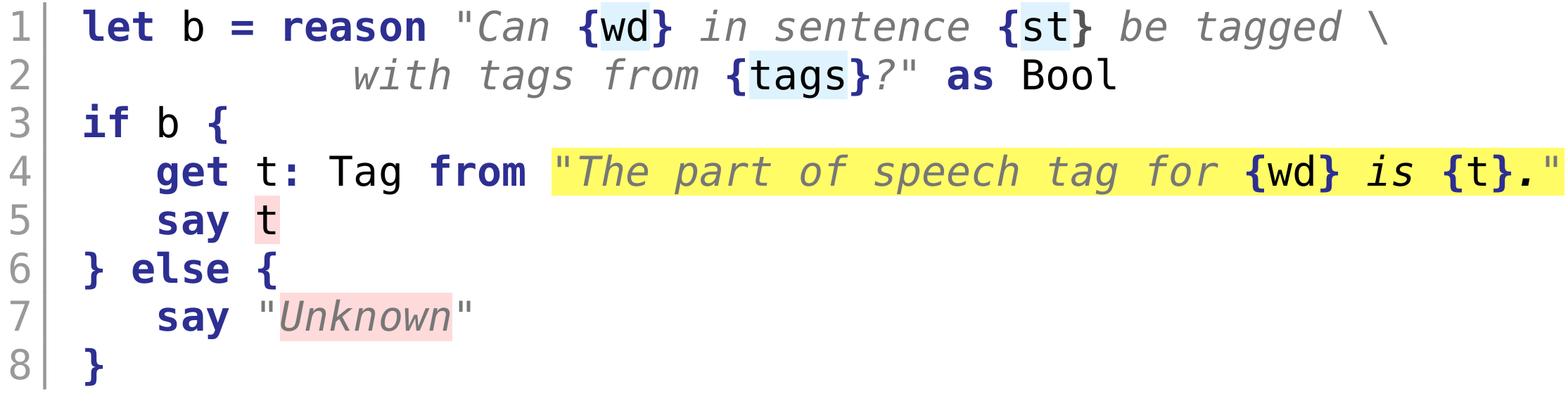}
      \caption{PoS in $\DSL$}\label{fig:pos:2}
    \end{subfigure}
    \begin{subfigure}{\textwidth}
        \centering
        \includegraphics[width=\linewidth]{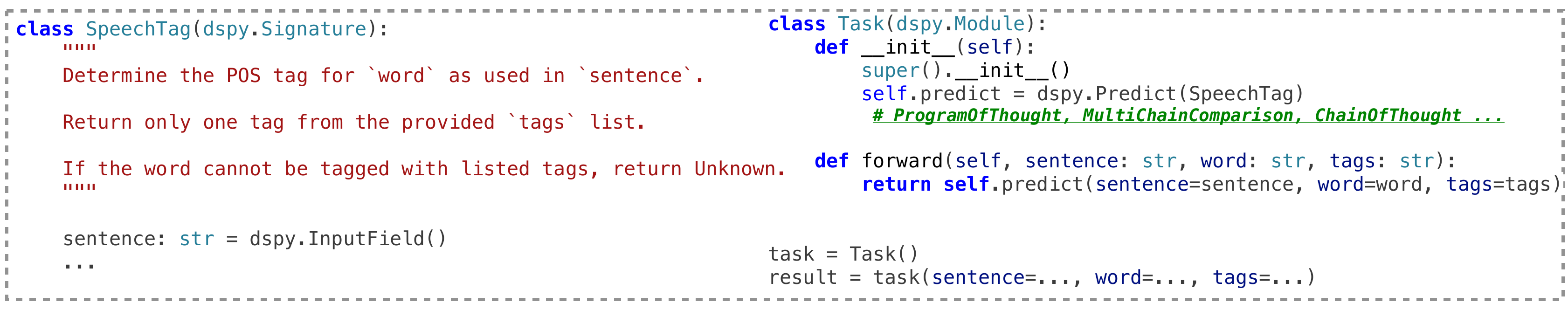}
        \caption{PoS in DSPy}\label{fig:pos:3}
    \end{subfigure}
    \caption{Motivating example: a Part-of-Speech (PoS) tagging prompt.~\cite{dspyhomepage}}
    \label{fig:pos}
\end{figure}

\paragraph{Prompt Description via $\DSL$}
\vspace{-1ex}
To address these challenges, 
we propose $\DSL$, a high-level domain-specific language for prompt description.
\Cref{fig:pos:1,fig:pos:2} show a simple part-of-speech (PoS) tagging prompt, 
adapted from PromptPex~\cite{promptpexsharma2025automatictestgeneration}, 
in natural language and in $\DSL$, respectively.
In the example, \sethlcolor{myBlue}\hl{\,\texttt{blue}\,} text denotes prompt inputs, 
\sethlcolor{myRed}\hl{\,\texttt{red}\,} text denotes expected outputs, 
and \sethlcolor{myYellow}\hl{\,\texttt{yellow}\,} text denotes the core task description.
The $\DSL$ program uses $\Reason$ to pose a task to the LM (line~\Line{1}), 
$\Get$ to retrieve a typed response and bind it to a variable (line~\Line{4}), 
and $\Say$ to emit output (lines~\Line{5}~and~\Line{7}).
Control flow is expressed with standard $\If/\Else$ conditionals.
The type system is a core feature of our language: user-defined types such as \texttt{Tag} constrain LM outputs to valid formats, so that structured extraction (e.g., line~\Line{4}) is both declarative and checkable.
Our naming and design are inspired by the classic pedagogical language $\IMP$~\cite{Winskel1993TheFS}.
Like $\IMP$, we aim to provide a minimal set of language constructs sufficient to express the core logic of prompt programs. We illustrate a detailed example and formally present the language in~\Cref{sec:lang} as our first contribution.

$\DSL$ addresses the first challenge, 
i.e., the coupling of task description with execution details, 
by explicitly decoupling program description and compilation strategy.
Users write their task as a structured program in $\DSL$ and use one of the predefined compilation strategies.
Each strategy handles low-level prompting details such as hyperparameter selection.
\Cref{fig:pos:3} shows how the program would look in DSPy.
The class \texttt{Task} requires users to manually specify the module type, 
with options including \texttt{dspy.Predict}, \texttt{dspy.ProgramOfThought}, \texttt{dspy.MultiChainComparison}, \texttt{dspy.ChainOfThought}, and more.
These are execution details, unrelated to the task description itself.
This comparison illustrates how $\DSL$ allows users to provide prompt descriptions without worrying about execution details. We identify this decoupling design principle as our second core contribution.

Turning to the second challenge, 
i.e., the lack of clear semantics for compilation strategies, 
our third contribution is to formalize two compilation strategies for $\DSL$ programs.
The first, which we call the \emph{baseline} strategy, 
compiles an $\DSL$ program into a one-shot prompt description and uses a static analysis to generate a global decoding constraint that constrains the LM output space.
The second, which we call the \emph{stepwise} strategy, 
compiles an $\DSL$ program into a sequence of LM-interaction instructions that enforce the program's execution order.
As part of the stepwise compilation, we define a minimal intermediate representation (IR) containing a set of low-level instructions that we believe can serve as a compilation target.
The formalization of these strategies clarifies (and in fact, \emph{defines}) how language-level mechanisms in $\DSL$ (e.g., type system, type casting, control flow) map to execution behavior under an LM.
These two strategies are illustrative rather than exhaustive:
one can define a multitude of other compilation strategies.
A user does not author a compiler for each program; 
they write a task description and select among predefined strategies,
which we envision being chosen automatically in the future.
We believe that our work provides the scaffolding for a more dedicated study of prompt compilation techniques for different prompting languages for future work.

\paragraph{Implementation and Evaluation}
\vspace{-0.5ex}
We implement $\DSL$ as an embedded DSL in Python, 
called $\DSLPy$.
We evaluate it on two case studies: 
statutory reasoning, where 114 distinct statute-level functions over 9 sections are exercised on 276 cases, 
and data labeling/transformation, 
where we study 5 programs with 67 tests.

\paragraph{Main Contributions}
\vspace{-1ex}
In summary, our core contributions are:
\begin{enumerate}[label={(\arabic*)}, leftmargin=25pt, topsep=1pt]
    \item $\DSL$, a declarative prompt programming language for specifying prompts with a minimal set of constructs for LM interaction, decoupling prompt descriptions from execution details (\Cref{sec:lang:2}).
    \item A type system for $\DSL$ to specify constraints and reason about them (\Cref{sec:lang:3}).
    \item Formal definitions of two compilation strategies for $\DSL$ programs:
    \begin{itemize}
      \item A \emph{baseline} strategy that compiles an $\DSL$ program into a one-shot prompt together with type-directed decoding constraints (\Cref{sec:baseline}).
      \item A \emph{stepwise} strategy that compiles an $\DSL$ program into $\DSLIR$, an intermediate representation with primitive operations for LM interaction (\Cref{sec:stepwise}).
    \end{itemize}
    \item $\DSLPy$, a Python implementation of $\DSL$ and its compilers (\Cref{sec:eval:1}).
    \item Two case studies demonstrating the expressiveness and practical utility of $\DSL$: statutory reasoning (\Cref{sec:eval:2:1}) and data labeling/transformation (\Cref{sec:eval:2:2}).
\end{enumerate}

\section{Background}\label{sec:bg}

Here we review some background on language models and constrained decoding.

\paragraph{Language Models}
\vspace{-1ex}
Given a base alphabet $\Sigma$ containing finite characters,
a \emph{string} is a sequence $\omega\in\Sigma^{*}$,
and \emph{completions} of $\omega$ are strings that start with $\omega$.
A \emph{token vocabulary} $V$ where $\Sigma\subseteq V\subseteq \Sigma^{*}$ is a set of finite strings of characters with a special token $\EOS$.
A \emph{tokenizer} $\Tok:\Sigma^{*}\to V^{*}$ is a function that maps strings to sequences of tokens, 
and a \emph{detokenizer} $\Detok:R\to\Sigma^{*}$ maps token sequences back to strings.
An autoregressive \emph{language model} $M:V^{*}\to{V}\to\Brack{0,1}$ is a function that takes \emph{prefix} $\t=t_1\cdots t_n\in V^{*}$ as input and returns a probability distribution $\Pr_M(\cdot\mid \t)$ over the next token $t_{n+1}\in V$.

\paragraph{Decoding}
\vspace{-1ex}
Fix a language model $M$ and a decoding policy,\footnote{We write only $M$ for simplicity. Formally, the induced response distribution also depends on decoding hyperparameters such as temperature, top-$p$, and any output-length limit.}
which together determine for each prefix $\t\in V^{*}$ a distribution $\Pr_M(\cdot\mid\t)$ over the next token.
Starting from the prompt prefix $\Tok(s)$, decoding repeatedly samples a token and appends it to the current prefix.
If the sampled tokens are $t_1,\ldots,t_k$, then the successive prefixes are
$\Tok(s)$, $\Tok(s)\cdot t_1$, $\Tok(s)\cdot t_1t_2$, and so on.
Decoding terminates when $\EOS\in V$ is generated.
We call any sequence in $R=(V\setminus\{\EOS\})^{*}\cdot\EOS$ a valid \emph{response}.
For each prompt string $s\in\Sigma^{*}$, decoding induces a distribution $\D[s]$ over $R$.
For any response $\t=t_1\cdots t_k\in R$,
\vspace{-3ex}
\[
\D[s](\t)=\prod_{i=1}^{k}\textstyle{\Pr_M}(t_i\mid\Tok(s)\cdot t_1\cdots t_{i-1}).
\vspace{-1ex}
\]
That is, $\D[s]$ is the probability distribution over finite token-sequence responses generated from $s$.

\paragraph{String constraint}
\vspace{-1ex}
A \emph{string constraint} is a decidable predicate $\varphi$ over $\Sigma^{*}$. 
We say a string $\omega\in\Sigma^{*}$ \emph{satisfies} $\varphi$ if and only if $\varphi(\omega)$ is true,
written as $\omega\models\varphi$,
and we define the set of all strings that satisfy $\varphi$ as $\Mod(\varphi)=\{\omega\in\Sigma^{*}\mid\omega\models\varphi\}$, 
called the \emph{model} of $\varphi$.
Two string constraints $\varphi$ and $\psi$ are \emph{equivalent}, 
written as $\varphi\equiv\psi$, if and only if $\Mod(\varphi)=\Mod(\psi)$.
In this paper we require $\varphi$ to be decidable, i.e., that there exists a decision procedure that determines whether any given string satisfies $\varphi$.
String constraints can be presented in many forms, including JSON, XML, regular grammars, 
and context-free grammars (CFG).

\paragraph{Constrained decoding}
\vspace{-1ex}
A \emph{constrained decoding} process takes a string constraint $\varphi$ as input and generates a response $\omega\in R$ such that $\Detok{\omega}\models\varphi$.
When the distribution and constraint are fixed, 
there are likewise many ways to decode, 
including beam search, sampling, and greedy decoding~\cite{hokamp-liu-2017-lexically,post-vilar-2018-fast,lu-etal-2021-neurologic}.
We abstract away from these details, and treat a string constraint as an abstract predicate and decoding as a function $\Decode{\D[s]}{\varphi}$ that takes a distribution $\D[s]$ and a string constraint $\varphi$ and returns a response $\omega\in R$. 

\paragraph{Notations}
\vspace{-1ex}
When we write string literals from $\Sigma^{*}$ in the meta-language, we use \Text{gray italic text}. 
We assume $\Sigma$ includes all symbols used in examples throughout this paper, 
including letters, digits and punctuation.

\section{The \texorpdfstring{$\DSL$}{Imprompt} Language}\label{sec:lang}

In this section, we first present the core language features of $\DSL$ in~\Cref{sec:lang:1} with an illustrative example. We then give the formal syntax in~\Cref{sec:lang:2} and the type system in~\Cref{sec:lang:3}.
Finally, we discuss the challenges of designing semantics for prompt languages and explain why the compilation semantics of $\DSL$ deserves careful analysis in~\Cref{sec:lang:4}.

\subsection{Illustrative Example}\label{sec:lang:1}

We demonstrate the key language features of $\DSL$ using the example in \Cref{fig:copyright}, which describes the procedure for analyzing whether a given use of copyrighted material constitutes Fair-Use. 
The intended task for the LM to solve can be described in English as follows:

{
\small 
\vspace{.5em}
\fontsize{9}{10}\selectfont
\begin{center}
    \begin{minipage}{0.9\textwidth}
        \Text{Assess if the use of copyrighted material in this work constitutes Fair-Use. First, determine if the new work is highly transformative. If it is, then assess its commercial impact. If the commercial impact on the original copyright holder is negligible, classify it as `Likely Fair Use', otherwise classify it as `Borderline Case'. If the work is not transformative, it is `Unlikely Fair Use' unless its purpose is clearly parody or critique.}
    \end{minipage}
\end{center}
}

\begin{figure}[htbp!]
\begin{lstlisting}[language=Imprompt]
let _ = reason "Assess if the use of given material constitutes Fair Use."  // task description
let x = "{INPUT}"                                           // external variable
let b = reason "Determine if {x} is highly transformative." as Bool  // type casting
if b {                                                      // control flow
    let imp = reason "Access the commercial impact of {x}."
    let neg = reason "Determine if {imp} on the original copyright holder is negligible." as Bool
    if neg {
        say "Likely Fair Use."                              // formatted output
    } else {
        say "Borderline Case."
    }
} else {
    let pur = reason "Determine if the purpose of {x} is clearly parody or critique." as Bool
    if !pur {                                               // boolean operation
        get res: Dyn from "Unlikely Fair Use, because {res}."  // constrained extraction
        say res
    }
}
\end{lstlisting}
    \vspace{1ex}
    \caption{A $\DSL$ program for Fair-Use Copyright Analysis.}
    \label{fig:copyright}
    \vspace{-2ex}
\end{figure}

\paragraph{Constructs with LM interactions}
The main distinction between prompt programming languages and classical programming languages is that the former includes primitives for interacting with LMs.
$\DSL$ has three such constructs.
\begin{enumerate}[label=(\arabic{*})]
    \item $\ReasonExpr{e}$ takes an expression $e$ as a step of task description and returns the LM response.
    For example, in line~\Line{1}, the overall task is described as \Text{"Assess if the use of given material constitutes Fair-Use."} and the return value is assigned to an unused variable. 
    In this example the response is discarded, but in general it may be bound to a variable and used in later steps of the program.
    This corresponds to one step of computation in a classical language. 
    \item $\CastExpr{e}{\tau}$ takes an expression $e$ as a prompt and asks the LM to interpret the response as a value of type $\tau$.
    For example, in line~\Line{3}, 
    we cast the $\Reason$ return value from the LM to type $\Bool$.
    This corresponds to type casting in a classical language, with the LM as the execution engine.
    We explain how such casting is implemented in~\Cref{sec:baseline}.
    \item $\GetStmt{x\Colon\tau}{t}$ asks the LM to fill in a hole in a template $t$ with a value of type $\tau$ and bind the filled value to a fresh variable $x$.
    For example, in line~\Line{15}, we ask the LM to fill in the hole in template \Text{"Unlikely Fair-Use, because \Brace{res}"} with a value of type $\Dyn$ explaining the reason, 
    which enforces no constraints, and bind the filled value to variable \texttt{res}.
    This corresponds to a function call in a classical language.
\end{enumerate}

\paragraph{Variables and environments}
\begin{wrapfigure}{r}{0.5\textwidth} %
    \vspace{-1em}
    \centering
    \includegraphics[width=0.5\textwidth]{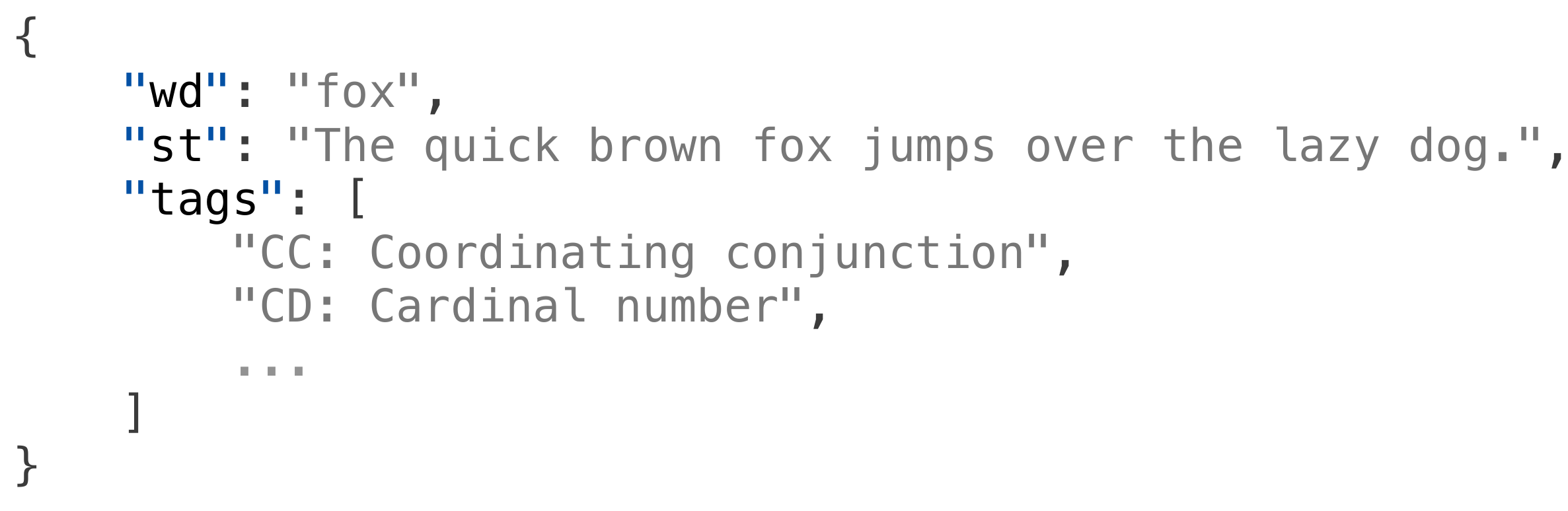}
    \caption{Global environment for~\Cref{fig:pos:2} in JSON}\label{fig:sigma}
    \vspace{-1em}
\end{wrapfigure}

$\DSL$ uses $\Let$ and $\Get$ to declare internal variables, 
and also supports external variables from a global environment immutable to $\DSL$ programs.
External variables are not declared in the program but may be referenced in templates (e.g., \texttt{INPUT} on line~\Line{2}).
This design reflects the fact that practical prompts are often not plain text, 
but rather \emph{prompt templates} with placeholder variables~\cite{10.1145/3696630.3728533}.
In practice, such an environment often stores input data and can be provided in JSON format (\Cref{fig:sigma}).

\paragraph{Types and type casting}
Although interactions with LM are text-based, 
$\DSL$ uses types to express constraints on LM responses and type casting to enforce them.
In particular, $\DSL$ provides built-in expression types $\Dyn$ and $\Bool$.
$\Dyn$ denotes unknown values produced at LM boundaries with no constraint, 
while $\Bool$ constrains LM responses to \Text{true} or \Text{false}.
For example, on line~\Line{3}, the LM response is cast to type $\Bool$ using the $\As$ construct.
The cast thus enables its use as a boolean guard at line~\Line{4}.
In addition to built-in types, users may define custom types $\Ty{\omega,\varphi}$, 
where $\omega \,{\in}\, \Sigma^{*}$ is a natural language description of this type and $\varphi$ is a string constraint giving the formal specification.
For example, in the PoS example in~\Cref{fig:pos:2}, \texttt{Tag} is a user-defined type with description $\omega \,{=}\, \Text{"a part-of-speech tag"}$ and a string constraint $\varphi$ that specifies the set of PoS tags (e.g., \texttt{\char`^(\Text{CC}|\Text{CD}|...)\textdollar} as a regex).
Custom types are treated as a meta-language construct in this paper since we formulate string constraints as abstract predicates. The concrete syntax and surface form of string constraints can be implemented in different ways. 
In our implementation in this paper, users define custom types as Python classes. 
We discuss this design in detail in~\Cref{sec:eval}.

Readers may notice that our types are entirely string-based.
For example, we do not have an $\Int$ type.
This design reflects our view that the LM is a completion machine and all our interactions with it are essentially strings.
Of course, we can define an $\Int$ type as $\Ty{\Text{"an integer"}, \varphi}$, where $\varphi=\texttt{\char`\^[+-]?\textbackslash{d}+\textdollar}$ is a regex that only accepts string literals.

\paragraph{Control flow}
\vspace{-0.5ex}
$\DSL$ has control-flow structures similar to $\IMP$ at the syntax level, 
including sequencing, conditionals, and loops.
The key difference is that $\IMP$'s underlying execution model is a RAM model 
which executes instructions in the appropriate order, 
whereas $\DSL$ programs are sent to an LM that generates text one token at a time.
Consequently, we cannot simply assume that individual commands in prompt programs will be executed in the correct order.
Thus, control-flow structures in $\DSL$ serve primarily to express user intent, and it is the \emph{compiler} that must ensure the correctness of an execution. 
\Cref{sec:baseline} and~\Cref{sec:stepwise} adopt two different compilation strategies for controlling the flow of program execution. 
The first treats the prompt as holistic and model state transitions as a black box, 
and the second uses an external stack to enforce execution order.

\paragraph{Formatted output}
\vspace{-0.5ex}
Finally, $\DSL$ uses $\SayStmt{e}$ to output results (e.g., lines~\Line{8},~\Line{10}, and~\Line{16}).
The implementation of $\SayStmt{e}$ statements depends entirely on the compilation strategy: 
the expression $e$ is evaluated to a string, 
and the compiler may either output it directly without invoking the LM or add it to the LM context to change subsequent distributions.
In~\Cref{sec:stepwise} we will elaborate on this in detail.

\subsection{Core Syntax}\label{sec:lang:2}

\Cref{fig:syntax} presents the core syntax of $\DSL$.
Like $\IMP$, we intentionally keep $\DSL$ small: 
it uses a minimal set of constructs sufficient to express structured prompting procedures. 
We focus here on the formal presentation of the syntax, 
and readers seeking intuition may refer to \Cref{sec:lang:1} for an introduction to the language constructs.

\begin{figure}[htbp!]
\vspace{-1ex}
    {\fontsize{9}{8}\selectfont
    \begin{align*}
        \Lit~l ::= ~~~ & ~s\in\Sigma^{*}~\mid~\True~\mid~\False \\
        \Tmpl~t ::= ~~~ & ~s\in\Sigma^{*}~\mid~\Brace{x}~\mid~{t_1}{t_2} \\
        \Type~\tau ::= ~~~ & \Dyn~\mid~\Bool~\mid~\Ty{\omega,\varphi}\\
        \Expr~e ::= ~~~ & ~x~\mid~\Text{"}l\Text{"}~\mid~\Text{"}t\Text{"}~\mid~\Not{e}~\mid~e_1~\And~e_2~\mid~e_1~\Or~e_2~\mid~\ReasonExpr{e}~\mid~\CastExpr{e}{\tau} \\
        \Stmt~c ::= ~~~ & ~\Skip~\mid~\LetEqStmt{x}{e}~\mid~\GetStmt{x\Colon\tau}{t}~\mid~\IfStmt{e}{c_1}~\ElseStmt{c_2} \\
        & ~\hspace{5pt}\mid~\WhileStmt{e}{c}~\mid~\SayStmt{e}~\mid~\SeqStmt{c_1}{c_2} 
        \end{align*}
    }
    \caption{$\DSL$ core syntax.}\label{fig:syntax}
    \vspace{-2ex}
\end{figure}

\paragraph{Literals}
\vspace{-1ex}
Fix a base alphabet $\Sigma$ of symbols,
$\Lit$ contains all strings over $\Sigma$ and boolean literals $\True$ and $\False$.

\paragraph{Templates}
Templates in $\DSL$ contain all concrete strings and may embed a variable $x$ as $\Brace{x}$,
which denotes \emph{literal substitution} of the value of $x$ from current scope into the prompt.
A template $t$ is well formed only if it contains at least one variable occurrence. We do not syntactically distinguish between internal and external variables.

\paragraph{Types}
Types in $\DSL$ include the dynamic type $\Dyn$, 
the boolean type $\Bool$, 
and user-defined types $\Ty{\omega,\varphi}$.
As we discussed in~\Cref{sec:lang:1}, 
every type $\tau$ gives rise to a string constraint $\C[\tau]$, 
which we formally define via their models as follows:
\[
\Mod\Paren{\C[\Dyn]}=\Sigma^{*}\quad
\Mod\Paren{\C[\Bool]}=\{\Text{true},\Text{false}\}\quad
\Mod\Paren{\C[\Ty{\omega,\varphi}]}=\Mod(\varphi).
\]
Two types are \emph{equivalent}, denoted by $\tau_1\equiv\tau_2$,
if they induce equivalent string constraints:
\[
\tau_1\equiv\tau_2 \overset{\triangle}{\iff}\C[\tau_1]\equiv\C[\tau_2].
\]

\paragraph{Expressions}
Expressions in $\DSL$ include variables, quoted string literals, quoted templates, boolean operations,
the reasoning expression $\ReasonExpr{e}$ and the type-casting expression $\CastExpr{e}{\tau}$.

\paragraph{Statements}
A statement $c$ is a computation step that produces no value but may have side effects.
A \emph{program} $P$ is a top-level statement.

\subsection{Type System}\label{sec:lang:3}

The type system of $\DSL$ assigns a type to an expression and determines whether a statement is well-formed. 
In~\Cref{fig:typing-rules-expr},
we define the typing judgments for expressions.
A context $\Gamma$ is a mapping from variables to types. 
The judgment $\Gamma\vdash e:\tau$ states that under context $\Gamma$, the expression $e$ is well-typed with type $\tau$.
In~\Cref{fig:typing-rules-stmt}, 
we define the judgment $\Gamma\vdash c\dashv\Gamma'$, 
stating that under $\Gamma$, 
statement $c$ is well-formed and yields new context $\Gamma'$.

\vspace{1ex}
{
{\fontsize{9}{8}\selectfont
\begin{mathparpagebreakable}
    \inferrule[T-Var]
    {\Gamma(x)=\tau}
    {\Gamma\vdash x:\tau}
    \quad
    \inferrule[T-Tmpl]
    {~}
    {\Gamma\vdash \Text{"}t\Text{"}:\Dyn}
    \quad
    \inferrule[T-Lit]
    {~}
    {\Gamma\vdash \Text{"}l\Text{"}:\Dyn}
    \quad
    \inferrule[T-Bool]
    {e\in\{\True,\False\}}
    {\Gamma\vdash e:\Bool}
    \quad
    \inferrule[T-Not]
    {\Gamma\vdash{e}:\Bool}
    {\Gamma\vdash\Not{e}:\Bool}
    \and
    \inferrule[T-Bop]
    {
        \Gamma\vdash e_1:\Bool\quad
        \Gamma\vdash e_2:\Bool\quad
        \oplus\in\{\And,\Or\}
    }
    {\Gamma\vdash{e_1}\oplus e_2:\Bool}
    \quad
    \inferrule[T-Reason]
    {\Gamma\vdash{e}:\tau}
    {\Gamma\vdash\ReasonExpr{e}:\Dyn}
    \quad
    \inferrule[T-Cast]
    {\Gamma\vdash e:\tau'}
    {\Gamma\vdash\CastExpr{e}{\tau}:\tau}
\end{mathparpagebreakable}
}
\captionof{figure}{Typing rules for $\DSL$ expressions}
\label{fig:typing-rules-expr}
}

{\fontsize{9}{8}\selectfont
\begin{mathparpagebreakable}
    \inferrule[T-Skip\label{rule:t-skip}]
    {~}
    {\Gamma\vdash\Skip\dashv\Gamma}
    \quad
    \inferrule[T-Let\label{rule:t-let}]
    {
        \Gamma\vdash{e}:\tau\quad
        x\notin\Dom{\Gamma}
    }
    {\Gamma\vdash\LetEqStmt{x}{e}\dashv\Gamma,x:\tau}
    \quad
    \inferrule[T-Let-2\label{rule:t-let-2}]
    {
        \Gamma\vdash{e}:\tau\quad
        \Gamma(x)\equiv\tau
    }
    {\Gamma\vdash\LetEqStmt{x}{e}\dashv\Gamma}
    \quad
    \inferrule[T-Get\label{rule:t-get}]
    {
        \Gamma\vdash{e}:\tau'\quad
        x\notin\Dom{\Gamma}
    }
    {\Gamma\vdash\GetStmt{x\Colon\tau}{t}\dashv\Gamma,x:\tau}
    \\
    \inferrule[T-Say\label{rule:t-say}]
    {\Gamma\vdash{e}:\tau}
    {\Gamma\vdash\SayStmt{e}\dashv\Gamma}
    \quad
    \inferrule[T-Seq\label{rule:t-seq}]
    {
        \Gamma\vdash{c_1}\dashv\Gamma_1\\\\
        \Gamma_1\vdash{c_2}\dashv\Gamma_2
    }
    {\Gamma\vdash\SeqStmt{c_1}{c_2}\dashv\Gamma_2}
    \quad
    \inferrule[T-If\label{rule:t-if}]
    {
        \Gamma\vdash{e}:\Bool\\\\
        \Gamma\vdash{c_1}\dashv\Gamma_1\quad
        \Gamma\vdash{c_2}\dashv\Gamma_2
    }
    {\Gamma\vdash\IfStmt{e}{c_1}~\ElseStmt{c_2}\dashv\Gamma}
    \quad
    \inferrule[T-While\label{rule:t-while}]
    {
        \Gamma\vdash{e}:\Bool\quad
        \Gamma\vdash{c}\dashv\Gamma_1
    }
    {\Gamma\vdash\WhileStmt{e}{c}\dashv\Gamma}
\end{mathparpagebreakable}
\captionof{figure}{Well-formation rules for statements in $\DSL$}\label{fig:typing-rules-stmt}
\vspace{1ex}
}

The type information obtained from $\DSL$'s type checking serves two purposes.
\One~The primary purpose is to assist compilation, which we elaborate in detail in~\Cref{sec:baseline}.
\Two~The secondary purpose is to help programmers describe their intentions more precisely, 
to increase the likelihood of correct program execution under the stochastic nature of LMs.

\begin{wrapfigure}{r}{0.4\textwidth} %
    \centering
    \includegraphics[width=0.4\textwidth]{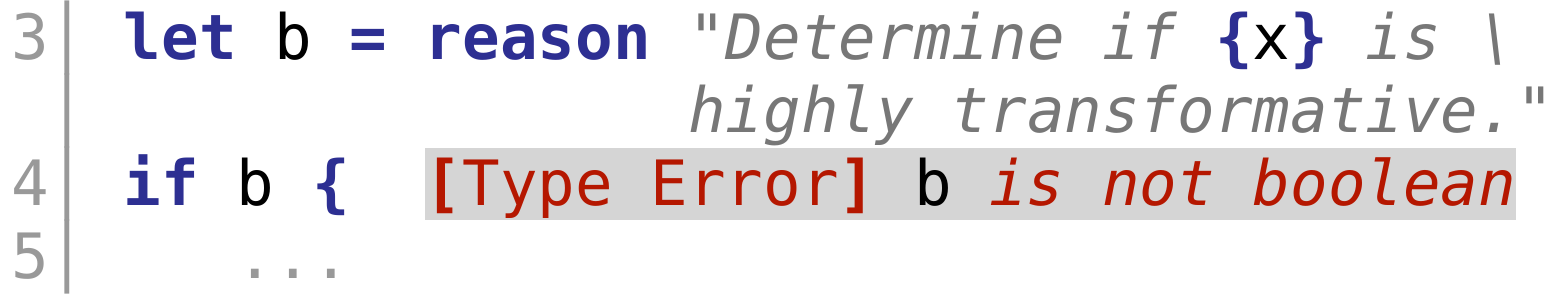}
    \caption{Types for precise intent description.}\label{fig:intent-dsl}
    \vspace{-2em}
\end{wrapfigure}
For example, 
\Cref{fig:intent-dsl} showcases the Fair-Use example from~\Cref{fig:copyright},
but with the modification that the user forgets to cast the LM response of line~\Line{3} to type $\Bool$. Line~\Line{4} will then raise a type error. 
With this type system, users are allowed and encouraged to provide more details about the prompt.
However, to preserve programming flexibility, 
such constraints are designed to be minimal.
Consequently, the overall type checking is relatively permissive.
The main source of this permissiveness is that users can cast any well-typed expression $e$ to any type $\tau$ by~\textsc{T-Cast}.
This differs from common programming languages and Gradual Typing, 
which typically require a compatibility relation $\lhd$ to ensure type safety for casts. Of course, such a relation could also be appropriate for $\DSL$. For instance, consider the following example:
\[
\LetEqStmt{x}{\Reason~\Text{"What day of the week is it today?"}~\As~\Bool}
\]
Here we cast the response to a general question to $\Text{true}$ or $\Text{false}$,
which is not only semantically questionable,
but may also skew the LM token distribution depending on the compilation strategy. However, we do not pursue this idea further in this work, and focus only on demonstrating that types provide a mechanism for constraining prompt programming. We leave a deeper exploration of typing for prompt programming languages to future work. 

Type checking for $\DSL$ is decidable. 
The judgments in~\Cref{fig:typing-rules-expr,fig:typing-rules-stmt} can be decided by a recursive procedure defined by structural recursion on the syntax of expressions and statements.
We denote the type checking procedure as $\Typeof{e}$ for a well-typed expression $e$.
Moreover, the type system of $\DSL$ ensures uniqueness up to type equivalence, as stated in the following lemma.
\begin{lemma}[Uniqueness]\label{lem:uniqueness}
If $\,\Gamma\vdash e:\tau$ and $\Gamma\vdash e:\tau'$, then $\tau\equiv\tau'$.
\begin{proof}
    By induction on the structure of $e$.
\end{proof}
\end{lemma}

\subsection{Discussion of Compilation Semantics}\label{sec:lang:4}

In~\Cref{sec:lang:1},
we explained the natural intent of each $\DSL$ construct with~\Cref{fig:copyright},
but we did not describe a singular formal semantics. 
This is because the prompt domain currently lacks a clear, widely-accepted set of low-level primitives that would allow us to describe LM behavior or state transitions (we attempt to remedy this in part in~\Cref{sec:stepwise:1}).

Classical programming languages are different in this respect.
For example, consider $\IMP$,
whose small-step operational semantics is often defined as a transition relation $\to~\subseteq(\Stmt\times\Store)^2$,
where $\Store:\Var\rightharpoonup\Int$ is a partial map from variables to integers.
The transition relation is defined in this way because it relies on the underlying RAM computation model as a standard semantic foundation,
whose operation set is familiar, finite, and discrete.
In contrast, in $\DSL$,
the semantics of the three constructs $\Reason$, $\Get$, 
and $\As$ do not have a simple formulation (such as a transition relation), 
because they depend on LM interactions.

\begin{figure}[htbp!]
    \centering
    \includegraphics[width=.85\textwidth]{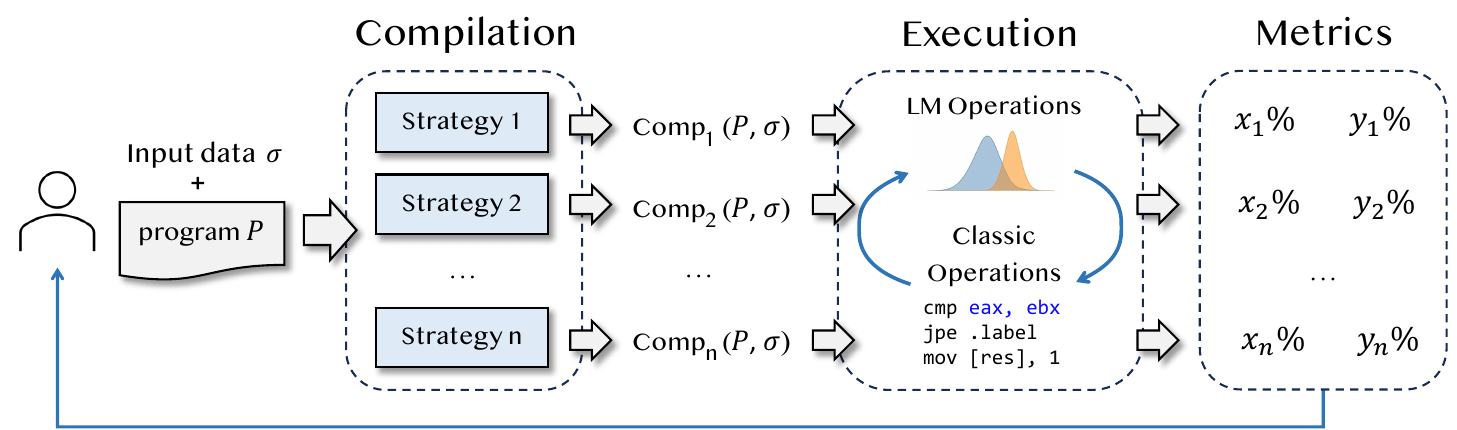}
    \vspace{2ex}
    \caption{Semantics via backward simulation.}
    \label{fig:comp-sem}
    \vspace{-2ex}
\end{figure}

In the prompt language setting, 
we take the \emph{backward simulation} approach as the most appropriate way to describe $\DSL$:
namely, given an $\DSL$ program $P$, input $\sigma$, and a compilation strategy $\Comp$,
any observable behavior $B$ of $\Comp(P,\sigma)$ \emph{is} an acceptable behavior of $(P,\sigma)$,
which we denote as
\[
\forall B.~(P,\sigma)\Downarrow B\implies\Comp(P,\sigma)\Downarrow B
\] 
In this sense, every compiler \emph{defines} a semantics for $\DSL$.
This view also raises a challenge: 
different compilers may be described in terms of different low-level primitives, 
making it difficult to reason about their \emph{equivalence} in a single uniform semantic framework.
We therefore leave a formal equivalence theory to future work, 
and instead evaluate compilers by metrics over observable behaviors $B$.
In this paper, $B$ is mainly related to the LM output.
\Cref{fig:comp-sem} illustrates this perspective: 
a prompt program $P$ is compiled into a sequence of operations, 
combining LM interactions with classic computation, 
and different compilations can be compared by their resulting behaviors.

In~\Cref{sec:baseline} and~\Cref{sec:stepwise}, 
we present two compilation strategies for $\DSL$. 
Each induces a distinct semantics for the language. 
In~\Cref{sec:eval}, we define evaluation metrics such as success rate and test coverage 
and compare the performance of the two strategies across different case studies. 
The compilation space of $\DSL$, however, is far from exhausted, 
and we discuss other possible compilation strategies in~\Cref{sec:discussion}.

\section{Baseline Compilation Strategy}\label{sec:baseline}

We present a \emph{baseline} compilation strategy for $\DSL$.
As reviewed in~\Cref{sec:bg}, 
each prompt string $s \,{\in}\, \Sigma^{*}$ induces a distribution $\D[s]$ over finite LM completions.
Intuitively, the baseline compiler takes a $\DSL$ program and input data, 
and produces a prompt description together with a string constraint.
We formalize it as follows.
\begin{definition}[Baseline compiler]\label{def:baseline-compiler}
    The baseline compiler is the function
    \[
    \BComp(P,\sigma)=(\D\Brack{\Sem{P}\sigma},\C[P]),
    \]
    where
    \begin{itemize}[leftmargin=2em]
    \item $P$ is a $\DSL$ program,
    \item $\sigma:\Var\to\Sigma^{*}$ is a global environment of external variables (i.e., the input data),
    \item $\Sem{P}$ is a textual prompt template obtained from $P$ via the transformation function $\Sem{\cdot}$,
    \item $\Sem{P}\sigma$ is the string after substituting $\sigma$ into $\Sem{P}$, and
    \item $\C[P]$ is a global type-directed string constraint over the LM response.
    \end{itemize}
\end{definition}

The core of this compiler is the transformation function
$\Sem{\cdot}$, which defines how to transform a $\DSL$ program into a structured prompt artifact.
We describe $\C[P]$ as \emph{type-directed} because 
types are the smallest units that express constraints in the program (\Cref{sec:lang:1}),
and we show in~\Cref{sec:baseline:2} that a static analysis suffices to derive $\C[P]$.
We define an \emph{execution} of the compiled program as a constrained decoding process of $\D\Brack{\Sem{P}\sigma}$ under the constraint $\C[P]$.

\begin{figure}[htbp!]
    \centering
    \includegraphics[width=\textwidth]{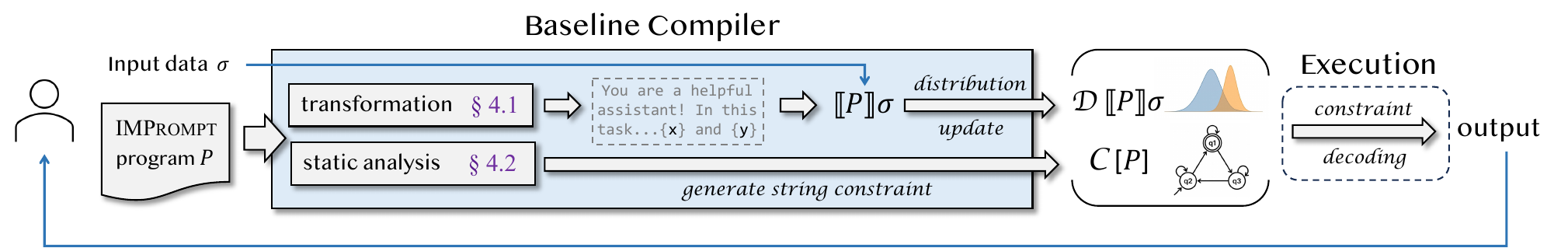}   
    \caption{Baseline compilation workflow}
    \label{fig:baseline}
\end{figure}

\Cref{fig:baseline} illustrates the overall workflow of baseline compilation and execution.
We dedicate the remainder of this section to the presentation of the transformation function (\Cref{sec:baseline:1})  and type-directed constraint $\C[P]$ (\Cref{sec:baseline:2}). 

\subsection{Compiling Programs to Prompt Templates}\label{sec:baseline:1}

The transformation function $\Sem{\cdot}$ for a $\DSL$ program is defined on~\One~types,~\Two~expressions, and~\Three~statements,
which we present in Figure~\ref{fig:baseline-type},\ref{fig:baseline-expr}, and~\ref{fig:baseline-stmt} respectively.
\begin{figure}[htbp!]
\centering
\vspace{-1em}
{\fontsize{9}{8}\selectfont
\begin{align*}
    \Sem{\Dyn} = \Text{a value}\qquad
    \Sem{\Bool}= \Text{a boolean value}\qquad
    \Sem{\Ty{\omega,\varphi}}=\omega
\end{align*}
}
\caption{$\Sem{\cdot}:\Type\to\Sigma^{*}$}
\label{fig:baseline-type}
\end{figure}

{
\centering
\vspace{-1em}
{\fontsize{9}{8}\selectfont
\begin{mathparpagebreakable}
    \vspace{-.5em}
    \Sem{s}=s
    \qquad
    \Sem{\Brace{x}}=\Text{<var name="}x\Text{"/>} 
    \qquad
    \Sem{t_1t_2}=\Sem{t_1}\Sem{t_2}
    \qquad
    \Sem{\True} =\Text{<bool>true</bool>}
    \\
    \vspace{-.5em}
    \Sem{\False} =\Text{<bool>false</bool>}
    \quad
    \Sem{\Not e} = \Sem{e}\Text{ is false}
    \quad
    \Sem{\ReasonExpr{e}}=\Text{<reason>}\Sem{e}\Text{</reason>}
    \\
    \vspace{-.5em}
    \Sem{e_1~\And~e_2} =\Sem{e_1}~\Text{and}~\Sem{e_2}
    \quad
    \Sem{e_1~\Or~e_2} = \Sem{e_1}~\Text{or}~\Sem{e_2}
    \quad
    \Sem{\CastExpr{e}{\tau}} =\Text{<convert format="}\Sem{\tau}\Text{">}\Sem{e}\Text{</convert>}
\end{mathparpagebreakable}
}
\captionof{figure}{$\Sem{\cdot}:\Expr\to\Sigma^{*}$}
\label{fig:baseline-expr}
}

\begin{figure}[htbp!]
    \vspace{-1em}
    {\fontsize{9}{8}\selectfont
    \begin{align*}
        \Sem{\Skip} =&~\epsilon\\
        \Sem{\LetEqStmt{x}{e}} =&~\Text{<set var="}x\Text{">}\Sem{e}\Text{</set>}\\
        \Sem{\GetStmt{x\Colon\tau}{t}} =&~\Text{<generate var="}x\Text{" format="}\Sem{\tau}\Text{">}\\
        &~\Text{<source>}\Sem{t}\Text{</source></generate>}\\
        \Sem{\IfStmt{e}{c_1}~\ElseStmt{c_2}}
        =&~\Text{<if><condition>}\Sem{e}\Text{</condition>}\\
        &~\Text{<then>}\Sem{c_1}\Text{</then><else>}\Sem{c_2}\Text{</else></if>}\\
        \Sem{\WhileStmt{e}{c}}
        =&~\Text{<while><condition>}\Sem{e}\Text{</condition><body>}\Sem{c}\Text{</body></while>}\\
        \Sem{\SayStmt{e}} =&~\Text{<output>}\Sem{e}\Text{</output>}\\
        \Sem{\SeqStmt{c_1}{c_2}} =&~\Sem{c_1}\Sem{c_2}
    \end{align*}
    }
    \caption{$\Sem{\cdot}:\Stmt\to\Sigma^{*}$}
    \label{fig:baseline-stmt}
\end{figure}

Note that expressions and statements need not be closed: free variables are compiled to named placeholders in the prompt artifact.
XML-like tags are used to help the LM distinguish different parts of the prompt artifact.
Such structured prompts tend to perform better on specific models according to contemporary records on best practices~\cite{anthropic2024xmlprompting}

\paragraph{\One~Transforming types}
The first step is defining $\Sem{\tau}$ for types $\tau$.
Each $\DSL$ type $\tau$ is transformed to a natural language description of the constraint in $\Sigma^{*}$, 
inheriting the discussion in~\Cref{sec:lang:1}.

\paragraph{\Two~Transforming expressions}
Expressions are fairly standard except type casting $\CastExpr{e}{\tau}$,
where we ask the LM to make this prediction and then reinterpret.
Readers may notice that a $\GetStmt{x\Colon\tau}{t}$ statement is semantically similar to the combination of $\Let$-$\Reason$-$\As$,
\begin{gather*}
\LetEqStmt{x}{\CastExpr{\ReasonExpr{\Text{<source>}\Sem{t}\Text{</source>}}}{\tau}}.
\end{gather*}
where the gray ${\color{gray}x}$ denotes the placeholder for the variable $x$.
However, they have a key distinction: 
a $\Get$-statement is intended to perform a single-step computation, directly outputting a result in the required format,
whereas the latter performs a two-step computation, first reasoning to obtain a result, then reinterpreting it to produce another result conforming to the format.
This single-step vs two-step distinction has been discussed in prior work~\cite{tam-etal-2024-speak}.
However, with the baseline compilation strategy these two approaches are essentially indistinguishable to the LM because baseline compilation makes a one-shot LM call. 
The distinction becomes clearer in the compilation strategy presented in~\Cref{sec:stepwise}.

\paragraph{\Three~Transforming statements}
Statements are transformed into structured XML-like commands that expose control flow and data flow directly to the LM.
For example, a branching statement is compiled into an \Text{<if>} node with explicit \Text{<condition>}, \Text{<then>}, and \Text{<else>} children,
rather than being paraphrased in natural language.

In summary, we trust the LM enough in the baseline compilation strategy that the compiled prompt is designed to be as descriptive and direct as possible.
However, the task description itself is informal, 
so the generated prompt may not always lead to the intended reasoning process, 
and the LM could respond with something different from what is specified in $\Say$ statements.
To formally constrain LM output, we use types in $\DSL$.
We describe this in the next subsection.

\subsection{Type-directed Constraints}\label{sec:baseline:2}

Now we present how the baseline compiler produces the second artifact of the compilation,
a global constraint $\C[P]$ for a program $P$.
Following the same structure as above, 
we generate $\C[P]$ in three steps:
constraints from~\One~types,~\Two~expressions, and~\Three~statements.

\paragraph{\One~Constraints from types}
Recall from~\Cref{sec:lang:2} that a type $\tau$ itself gives rise to a formal string constraint $\C[\tau]$ on LM output,
which can be described via the set of its models
{\fontsize{9}{8}\selectfont
\[
\Mod\Paren{\C[\Dyn]}=\Sigma^{*},\qquad
\Mod\Paren{\C[\Bool]}=\{\Text{true},\Text{false}\},\qquad
\Mod\Paren{\C[\Ty{\omega,\varphi}]}=\Mod(\varphi).
\]
}
\begin{wrapfigure}{r}{0.45\textwidth} 
\vspace{-3ex}
    \centering
    \includegraphics[width=0.45\textwidth]{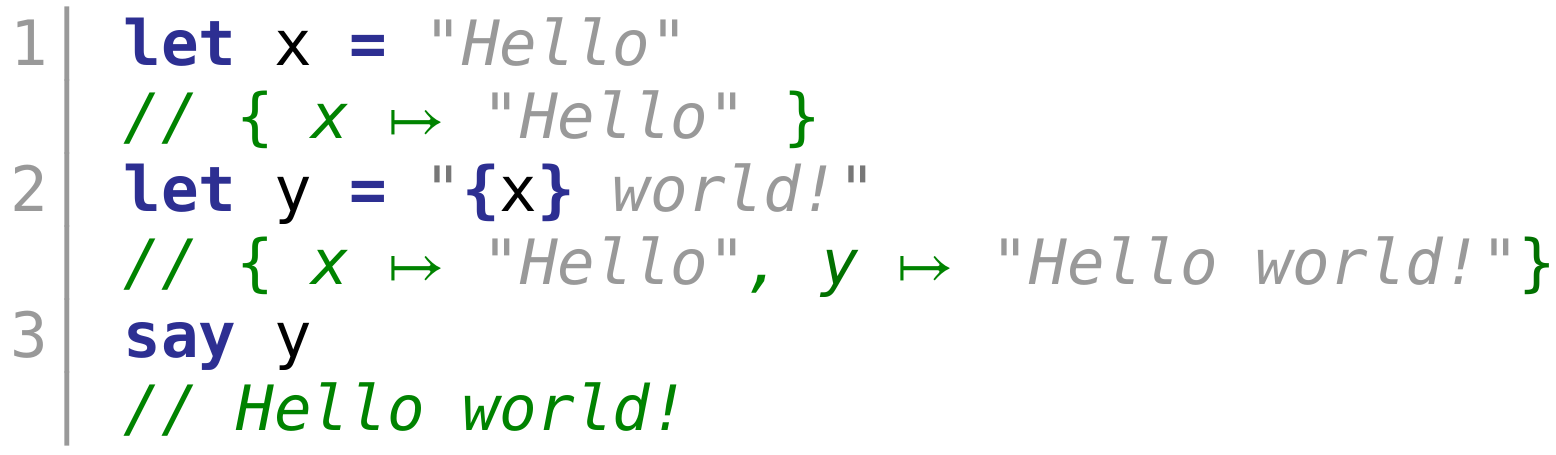}
    \caption{$\Gamma$ is too coarse for constraints}\label{fig:why-not-gamma}
    \vspace{-2em}
\end{wrapfigure}
At a high level, the type constraint is computed by statically modeling the execution of programs and iteratively replacing all occurrences of variables to their tightest constraints. 
However, using only $\Typeof{\cdot}$ would be too coarse for an informative constraint,
because our type system is designed to be minimal.
For example in~\Cref{fig:why-not-gamma},
if we use $\Typeof{\texttt{y}}$, which is $\Dyn$, 
the tightest constraint we know about the LM response on line~\Line{3} is $\C[\Dyn]$, 
which is essentially no constraint.
But obviously the tightest constraint should be the predicate that accepts only the string \Text{Hello world!}.
Thus, we model a finer-grained map $\Delta$ from variables to string constraints. 
For each variable $x$, 
$\Delta(x)$ records a constraint that the value of $x$ must satisfy:
\begin{enumerate}[label={(\arabic*)}, leftmargin=25pt, topsep=2pt]
    \item When $x$ is known to be $r\in\Sigma^{*}$, we record a singleton constraint $\psi$ such that $\{r\}=\Mod(\psi)$.
    \item When $x$ is only known up to type $\tau$, we record $\C[\tau]$.
    \item When $x$ is known up to a template $t$ with variables $x_1,\cdots,x_n$, where each $x_i$ is known up to constraint $\varphi_i$, we record a template constraint $\varphi_t$ such that
    \[\Mod(\varphi_t)=\Mod(t[x_1\mapsto\varphi_1,\cdots,x_n\mapsto\varphi_n])
    \]
\end{enumerate}
To combine constraints from sub-expressions and sub-statements, we need concatenation ($\varphi_1\cdot\varphi_2$), disjunction ($\varphi_1\lor\varphi_2$), and Kleene star ($\varphi^{*}$), defined via their models:
{\fontsize{9}{8}\selectfont
\begin{align*}
\Mod(\varphi_1\cdot\varphi_2)&\triangleq\Mod(\varphi_1)\cdot\Mod(\varphi_2)\\
\Mod(\varphi_1\lor\varphi_2)&\triangleq\Mod(\varphi_1)\cup\Mod(\varphi_2)\\
\Mod(\varphi^{*})&\triangleq\Mod(\varphi)^{*}
\end{align*}
}
We write $\bot$ for the constraint that rejects every string (used for statements that produce no output).
For control-flow merge points, we \emph{join} constraint environments pointwise, 
keeping the most precise information expressible in our constraint domain:
{\fontsize{9}{8}\selectfont
\[
(\Delta_1\sqcup\Delta_2)\triangleq
x\mapsto\begin{cases}
\Delta_1(x)\lor\Delta_2(x) & x\in\Dom{\Delta_1}\cap\Dom{\Delta_2},\\
\Delta_1(x) & x\in\Dom{\Delta_1}\setminus\Dom{\Delta_2},\\
\Delta_2(x) & x\in\Dom{\Delta_2}\setminus\Dom{\Delta_1}.
\end{cases}
\]
}
If two branches assign the same exact value to a variable, the join remains exact.
If they assign different values, the join records the disjunction of those possibilities.

\paragraph{\Two~Constraints from expressions}
The judgement $\Delta\vdash e\Leadsto\varphi$ states that under constraint environment $\Delta$, 
expression $e$ yields a constraint $\varphi$.
\Cref{fig:baseline-expr-constraint} shows the rules for generating constraints from expressions

{
{\fontsize{8}{8}\selectfont
\begin{mathparpagebreakable}
    \inferrule[E-Str\label{rule:e-str}]
    {   
        s\in\Sigma^{*}\quad
        \{s\} = \Mod(\varphi)
    }
    {\Delta\vdash{s}\Leadsto\varphi}
    \quad
    \inferrule[E-Var\label{rule:e-var}]
    {
        x\in\Dom{\Delta}
    }
    {\Delta\vdash\Brace{x}\Leadsto\Delta(x)}
    \quad
    \inferrule[E-Var-Ext\label{rule:e-var-ext}]
    {
        \{\sigma(x)\}=\Mod(\varphi)\\\\
        x\in\Dom{\sigma}\setminus\Dom{\Delta}
    }
    {\Delta\vdash\Brace{x}\Leadsto\sigma(x)}
    \quad
    \inferrule[E-Tmpl\label{rule:e-tmpl}]
    {
        \Delta\vdash{t_1}\Leadsto\varphi_1\quad
        \Delta\vdash{t_2}\Leadsto\varphi_2
    }
    {\Delta\vdash{t_1t_2}\Leadsto\varphi_1\cdot\varphi_2}
    \quad
    \inferrule[E-True\label{rule:e-true}]
    {
        \{\Text{true}\}=\Mod(\varphi)
    }
    {\Delta\vdash{\True}\Leadsto\varphi}
    \\
    \inferrule[E-False\label{rule:e-false}]
    {
        \{\Text{false}\}=\Mod(\varphi)
    }
    {\Delta\vdash{\False}\Leadsto\varphi}
    \;\;
    \inferrule[E-BOp\label{rule:e-bop}]
    {\oplus\in\{\And,\Or\}}
    {\Delta\vdash{e_1}\oplus{e_2}\Leadsto\C[\Bool]}
    \;\;
    \inferrule[E-Not\label{rule:e-not}]
    {~}
    {\Delta\vdash\Not{e}\Leadsto\C[\Bool]}
    \;\;
    \inferrule[E-Reason\label{rule:e-reason}]
    {~}
    {\Delta\vdash\ReasonExpr{e}\Leadsto\C[\Dyn]}
    \;\;
    \inferrule[E-Cast\label{rule:e-cast}]
    {~}
    {\Delta\vdash\CastExpr{e}{\tau}\Leadsto\C[\tau]}
\end{mathparpagebreakable}
}
\captionof{figure}{Generating constraints from expressions}\label{fig:baseline-expr-constraint}
}

\vspace{1ex}
These rules are intuitive.
\Rule{rule:e-str}{E-Str} and \Rule{rule:e-tmpl}{E-Tmpl} specify that string parts of templates are constrained precisely.
\Rule{rule:e-var}{E-Var} and~\Rule{rule:e-var-ext}{E-Var-Ext} indicate that we first try to retrieve a variable's constraint from $\Delta$,
and if it is not found there, 
we retrieve it from the global environment $\sigma$.
\Rule{rule:e-bop}{E-BOp} and~\Rule{rule:e-not}{E-Not} specify that the constraint for boolean expressions can be precise at most to $\C[\Bool]$.
\Rule{rule:e-reason}{E-Reason} and~\Rule{rule:e-cast}{E-Cast} specify that the constraints for $\ReasonExpr{e}$ and $\CastExpr{e}{\tau}$ can be precise at most to $\C[\Dyn]$ and $\C[\tau]$, respectively.

\paragraph{\Three~Constraints from statements}

The judgement $\Delta\vdash c\Leadsto\varphi\dashv\Delta'$ states that under constraint environment $\Delta$, statement $c$ yields a constraint $\varphi$ and an updated constraint environment $\Delta'$.
\Cref{fig:baseline-stmt-constraint} shows the rules for generating constraints from statements.

{\fontsize{8}{8}\selectfont
\begin{mathparpagebreakable}
    \inferrule[C-Skip\label{rule:c-skip}]
    {~}
    {\Delta\vdash\Skip\Leadsto\bot\dashv\Delta}
    \and
    \inferrule[C-Let\label{rule:c-let}]
    {
        \Delta\vdash{e}\Leadsto\varphi
    }
    {
        \Delta\vdash\LetEqStmt{x}{e}
        \Leadsto\bot\dashv\Delta[x\mapsto\varphi]
    }
    \and
    \inferrule[C-Get\label{rule:c-get}]
    {~}
    {
        \Delta\vdash\GetStmt{x\Colon\tau}{t}
        \Leadsto\bot\dashv\Delta[x\mapsto\C[\tau]]
    }
    \\
    \vspace{-.5em}
    \inferrule[C-Say\label{rule:c-say}]
    {\Delta\vdash e\Leadsto\varphi}
    {
        \Delta\vdash\SayStmt{e}
        \Leadsto\varphi\dashv\Delta
    }
    \quad
    \inferrule[C-Seq\label{rule:c-seq}]
    {
        \Delta\vdash{c_1}\Leadsto\varphi_1\dashv\Delta_1\quad
        \Delta_1\vdash{c_2}\Leadsto\varphi_2\dashv\Delta_2
    }
    {
        \Delta\vdash\SeqStmt{c_1}{c_2}
        \Leadsto\varphi_1\cdot\varphi_2\dashv\Delta_2
    }
    \quad
    \inferrule[C-If\label{rule:c-if}]
    {
        \Delta\vdash{e}\Leadsto\C[\Bool]\\\\
        \Delta\vdash{c_1}\Leadsto\varphi_1\dashv\Delta_1\quad
        \Delta\vdash{c_2}\Leadsto\varphi_2\dashv\Delta_2
    }
    {
        \Delta\vdash\IfStmt{e}{c_1}~\ElseStmt{c_2}
        \Leadsto\varphi_1\lor\varphi_2\dashv\Delta_1\sqcup\Delta_2
    }
    \\
    \inferrule[C-While\label{rule:c-while}]
    {
        \Floor{\Delta}\triangleq\Delta[x_i\mapsto\C[\Typeof{x_i}]]~\text{for all}~x_i\in\Dom{\Delta}
        \\\\
        \Floor{\Delta}\vdash{c}\Leadsto\varphi\dashv\Delta'
        \quad
        \Ceil{\Delta}\triangleq\Delta[x_i\mapsto\C[\Typeof{x_i}]]~\text{for all}~x_i\in\MV{c} 
    }
    {
        \Delta\vdash\WhileStmt{e}{c}
        \Leadsto\varphi^{*}\dashv\Ceil{\Delta}
    }
\end{mathparpagebreakable}
\captionof{figure}{Generating constraints from statements}
\label{fig:baseline-stmt-constraint}
}

\vspace{1ex}
By rules~\Rule{rule:c-skip}{C-Skip}, 
\Rule{rule:c-let}{C-Let}, and \Rule{rule:c-get}{C-Get},
$\Skip$, $\Let$, and $\Get$ impose no direct constraint on the output.
In $\LetEqStmt{x}{e}$,
the constraint on variable $x$ is updated to the constraint generated by $e$,
and in $\GetStmt{x\Colon\tau}{t}$,
the tightest constraint we can infer for $x$ is $\C[\tau]$, which is the constraint given by its type.
\Rule{rule:c-say}{C-Say} lifts the constraint of the expression to the statement level.
\Rule{rule:c-seq}{C-Seq} specifies that the constraint for a sequential statement is the concatenation of the constraints of the two sub-statements.
\Rule{rule:c-if}{C-If} specifies that the constraint for a conditional is the union of the constraints of its two branches, and that the post-state is the join of the two branch environments.

\Rule{rule:c-while}{C-While} widens the constraint environment to forget all information about modified variables but their types, 
and generate a new constraint that is the Kleene star of the widened body constraint, 
which is a sound approximation of the effect of any number of iterations.
$\MV{c}$ is the set of modified variables in $c$, as defined in~\Cref{fig:modified-vars}.

\begin{figure}[ht]
\vspace{2ex}
{\fontsize{9}{8}\selectfont
\begin{mathparpagebreakable}
\vspace{-.8em}
\MV{\Skip}=\emptyset\qquad
\MV{\SeqStmt{c_1}{c_2}}=\MV{c_1}\cup\MV{c_2}
\\
\vspace{-.8em}
\MV{\LetEqStmt{x}{e}}=\{x\}\quad
\MV{\GetStmt{x\Colon\tau}{t}}=\{x\}
\\
\vspace{-.8em}
\MV{\SayStmt{e}}=\emptyset\qquad
\MV{\WhileStmt{e}{c}}=\MV{c}
\\
\MV{\IfStmt{e}{c_1}~\ElseStmt{c_2} }=\MV{c_1}\cup\MV{c_2}
\end{mathparpagebreakable}
}
\caption{Definition of modified variables}\label{fig:modified-vars}
\vspace{-2ex}
\end{figure}

\smallskip
With~\Cref{fig:baseline-stmt-constraint},
we can generate a global constraint for a program $P$ by starting with an empty constraint environment and applying the rules to the top-level statement of $P$.
If $\emptyset\vdash{P}\Leadsto\varphi\dashv\Delta$,
then $\C[P]$ is defined as $\varphi$.
We conclude this section with two key properties of the generated constraints: well-definedness and soundness.

\begin{theorem}[Constraint Well-definedness]\label{thm:constraint-well-definedness}
$\C[P]$ is a well-defined string constraint for a $\DSL$ program $P$.
\begin{proof}[Proof Sketch]
    By the closure property of decidable languages under union, concatenation and Kleene star.
\end{proof}
\end{theorem}

\begin{theorem}[Constraint Soundness]\label{thm:constraint-soundness}
If an execution of the compiled program $\BComp(P,\sigma)$ produces a string $s$, 
then $s\models\C[P]$.
\begin{proof}[Proof Sketch]
    By Lemma~\ref{lem:uniqueness} and induction on the structure of $P$.
\end{proof}
\end{theorem}

\section{Stepwise Compilation Strategy}\label{sec:stepwise}

\begin{figure}[htbp!]
    \centering
    \includegraphics[width=\textwidth]{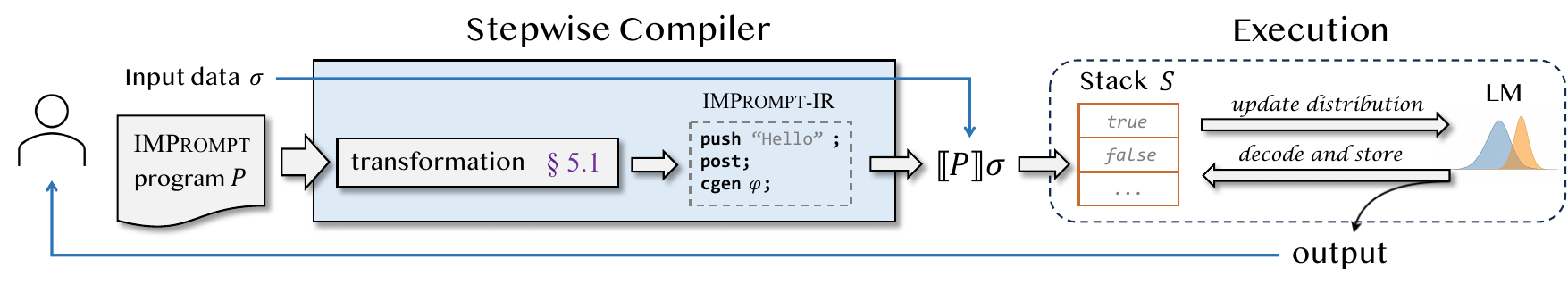}   
    \caption{Stepwise compilation workflow}
    \label{fig:stepwise}
    \vspace{-1ex}
\end{figure}

While the baseline compilation in~\Cref{sec:baseline} compiles $\DSL$ programs to prompt descriptions, 
it does not account for the \emph{temporal} nature of multi-turn prompting. 
To model the precise execution order,
we define a \emph{stepwise compilation} that compiles an $\DSL$ program into an initial state of a labeled transition system,
and define the execution as a sequence of transitions in this system.
\Cref{fig:stepwise} illustrates the overall workflow of stepwise compilation and execution.

We defer the formal definition of the stepwise compiler $\SComp$ to Def.~\ref{def:stepwise} after we introduce the intermediate representation and operational semantics in~\Cref{sec:stepwise:1} and~\Cref{sec:stepwise:2}.

\subsection{An Intermediate Representation}\label{sec:stepwise:1}

For convenience, 
we introduce $\DSLIR$,
an intermediate representation (IR) that makes LM interactions explicit.
The syntax of $\DSLIR$ is given in~\Cref{fig:stepwise-ir}.
\begin{figure}[htbp!]
\[
\begin{array}{rrl}
    \IR & ir ::= & \Post~\mid~\Gen~\mid~\Cgen{\tau}~\mid~\Push{s}~\mid~\Lookup{x}\\
    & \mid & \Pop{x}~\mid~\Dup~\mid~\Print~\mid~\Cat~\mid~\TemplateIR{t}~\mid~\AndOp~\mid~\OrOp~\mid~\NotOp \\
    & \mid & \Seq{ir}{ir}~\mid~\Branch{ir}{ir}~\mid~\IRWhile{ir}{ir}~\mid~\Nop
\end{array}
\]
\caption{Intermediate representation for stepwise compilation.}
\vspace{-2ex}
\label{fig:stepwise-ir}
\end{figure}

$\DSLIR$ operates on a stack $S$ of strings.
The three LM-specific primitives are $\Post$, $\Gen$, and $\Cgen{\tau}$.
$\Post$ pops the top of $S$ and sends it to the LM as the prompt.
$\Gen$ invokes the LM for an unconstrained completion and pushes the result onto $S$.
$\Cgen{\tau}$ invokes constrained decoding under the type-directed constraint $\C[\tau]$ and pushes the result onto $S$.

The remaining primitives are intuitive and familiar from conventional programming languages.
$\Push{s}$ pushes string $s$ onto $S$.
$\Lookup{x}$ looks up variable $x$ from a given environment and pushes its value onto $S$.
$\Pop{x}$ pops the top of $S$ and binds it to $x$.
$\Dup$ duplicates the top of $S$.
$\Print$ outputs the string on top of $S$.
$\TemplateIR{t}$ renders template $t$ against the current environment,
leaving any placeholders that are not substituted unchanged,
and pushes the resulting string onto $S$.
$\Seq{ir_1}{ir_2}$ sequences $ir_1$ and $ir_2$.
$\Cat$ pops the top two strings from $S$, 
concatenates them, and pushes the result.
$\AndOp$, $\OrOp$, and $\NotOp$ pop one or two strings, 
compute the logical operation, and push the result.
$\Branch{c_1}{c_2}$ pops the controlling string from the top of $S$ and executes $c_1$ if the value is \Text{true}, otherwise executes $c_2$.
$\IRWhile{c_1}{c_2}$ repeatedly runs $c_1$, 
leaving a boolean on $S$, pops it, and if true runs $c_2$ and repeats, 
otherwise stops.

Now we define a transformation function $\Sem{\cdot}$ from $\DSL$ statements and expressions to $\DSLIR$ instructions.
Rules for expressions are given in~\Cref{fig:stepwise-expr},
where $\Sem{e}$ produces a sequence of $\DSLIR$ instructions that evaluates $e$ and leaves its value (as a string) on $S$.
Rules for statements are given in~\Cref{fig:stepwise-stmt},
where $\Sem{c}$ produces $\DSLIR$ instructions that executes $c$.

{
\centering
{\fontsize{9}{8}\selectfont
\begin{mathparpagebreakable}
    \vspace{-.5em}
    \Sem{\Brace{x}} = \Lookup{x}
    \quad
    \Sem{t_1 t_2} = \Seq{\Sem{t_1}}{\Seq{\Sem{t_2}}{\Cat}}
    \quad
    \Sem{\True} = \Push{\Text{true}}
    \quad
    \Sem{\False} = \Push{\Text{false}}
    \\
    \vspace{-.5em}
    \Sem{\Not e} = \Seq{\Sem{e}}{\NotOp}
    \quad
    \Sem{e_1~\Or~e_2} = \Seq{\Sem{e_1}}{\Seq{\Sem{e_2}}{\OrOp}} 
    \quad
    \Sem{e_1~\And~e_2} = \Seq{\Sem{e_1}}{\Seq{\Sem{e_2}}{\AndOp}}
    \\
    \Sem{s}=\Push{s}
    \quad
    \Sem{\ReasonExpr{e}} = \Seq{\Sem{e}}{\Seq{\Post}{\Gen}}
    \quad
    \Sem{\CastExpr{\ReasonExpr{e}}{\tau}} = \Seq{\Sem{e}}{\Seq{\Post}{\Cgen{\tau}}}
\end{mathparpagebreakable}
}
\captionof{figure}{$\Sem{\cdot}:\Expr\to\IR$ for expressions.}
\label{fig:stepwise-expr}
}

\begin{figure}[htbp!]
    \centering
        {\fontsize{9}{8}\selectfont
        \begin{align*}
            \Sem{\LetEqStmt{x}{e}} &= \Seq{\Sem{e}}{\Pop{x}} \\
            \Sem{\GetStmt{x\Colon\tau}{t}} &= \Seq{\TemplateIR{t}}{\Seq{\Post}{\Seq{\Cgen{\tau}}{\Pop{x}}}} \\
            \Sem{\SayStmt{e}} &= \Seq{\Sem{e}}{\Print} \\
            \Sem{\IfStmt{e}{c_1}~\ElseStmt{c_2}} &= \Seq{\Sem{e}}{\Branch{\Sem{c_1}}{\Sem{c_2}}} \\
            \Sem{\WhileStmt{e}{c}} &= \IRWhile{\Sem{e}}{\Sem{c}} \\
            \Sem{\SeqStmt{c_1}{c_2}} &= \Seq{\Sem{c_1}}{\Sem{c_2}}
        \end{align*}
        }
    \caption{$\Sem{\cdot}:\Stmt\to\IR$ for statements.}
    \label{fig:stepwise-stmt}
    \vspace{-1ex}
\end{figure}

The rules in~\Cref{fig:stepwise-expr} and~\Cref{fig:stepwise-stmt} show that the $\DSLIR$ in~\Cref{fig:stepwise-ir} is indeed lower-level than $\DSL$,
since every $\DSL$ statement can be represented by some $\DSLIR$ instructions.
For example, $\CastExpr{\ReasonExpr{e}}{\tau}$ is compiled to a sequence of three instructions:
first evaluate $e$ and push the resulting question onto the stack,
then send this question to the LM as the current prompt,
and finally perform constrained decoding under $\C[\tau]$ and push the result onto the stack.

\subsection{Operational Semantics}\label{sec:stepwise:2}

Now we present the operational semantics for executing this $\DSLIR$ with a labeled transition system.
A transition is of the form
\vspace{-1ex}
\[
\Angle{\rho,\pi,S,ir}\xrightarrow{\ell}\Angle{\rho',\pi',S',ir'}
\]
from two states $\Angle{\rho, \pi, S, ir},\Angle{\rho', \pi', S', ir'}$ with label $\ell\in\Sigma^{*}$,
where $\rho,\rho':\Var\to\Sigma^{*}$ are the local environments,
$\pi,\pi':\Sigma^{*}$ are the context strings that record the prompt history,
$S,S'$ are the stacks of values from $\Sigma^{*}$,
$ir,ir'$ are current $\DSLIR$ instructions,
and $\ell\in\Sigma^{*}$ is the string that was printed during this transition.
The value stack $S$ has the top at the head,
and we write $\ell::S$ for pushing $\ell$ onto $S$.
Same as in~\Cref{sec:baseline},
we need a global environment $\sigma:\Var\to\Sigma^{*}$ provided for external variables.
Variable lookup uses $\rho$ first, 
then $\sigma$ if $x\notin\Dom{\rho}$.

The core transition rules are presented in~\Cref{fig:stepwise-sem}, 
which gives a direct formalization of the intuitive explanation in the previous section.
While most rules are standard, \Rule{rule:i-post}{I-Post}, \Rule{rule:i-gen}{I-Gen}, and \Rule{rule:i-cgen}{I-Cgen} are the rules that interact with the LM, and we elaborate on them below.

\vspace{1ex}
{\fontsize{8}{8}\selectfont
\begin{mathparpagebreakable}
\inferrule[I-Post\label{rule:i-post}]
{S = s::S'}
{\Angle{\rho,\pi,S,\Post}
\xrightarrow{\epsilon}
\Angle{\rho,\pi\cdot s,S',\Nop}}
\quad
\inferrule[I-Gen\label{rule:i-gen}]
{s = \Decode{\D[\pi]}{\top}}
{\Angle{\rho,\pi,S,\Gen}\xrightarrow{\epsilon}\Angle{\rho,\pi,s::S,\Nop}}
\quad
\inferrule[I-Cgen\label{rule:i-cgen}]
{s = \Decode{\D[\pi]}{\C[\tau]}}
{\Angle{\rho,\pi,S,\Cgen{\tau}}\xrightarrow{\epsilon}\Angle{\rho,\pi,s::S,\Nop}}
\\
\inferrule[I-Push\label{rule:i-push}]
{s\in\Sigma^{*}}
{\Angle{\rho,\pi,S,\Push{s}}\xrightarrow{\epsilon}\Angle{\rho,\pi,s::S,\Nop}}
\quad
\inferrule[I-Lookup\label{rule:i-lookup}]
{s = \rho(x)~\text{if}~x\in\Dom{\rho},~\text{else}~\sigma(x)}
{\Angle{\rho,\pi,S,\Lookup{x}}\xrightarrow{\epsilon}\Angle{\rho,\pi,s::S,\Nop}}
\quad
\inferrule[I-Pop\label{rule:i-pop}]
{S = s::S'\quad\rho'=\rho[x\mapsto s]}
{\Angle{\rho,\pi,S,\Pop{x}}\xrightarrow{\epsilon}\Angle{\rho',\pi,S',\Nop}}
\\
\inferrule[I-Dup\label{rule:i-dup}]
{S = s::S'}
{\Angle{\rho,\pi,S,\Dup}\xrightarrow{\epsilon}\Angle{\rho,\pi,s::s::S',\Nop}}
\quad
\inferrule[I-Print\label{rule:i-print}]
{S = s::S'}
{\Angle{\rho,\pi,S,\Print}\xrightarrow{s}\Angle{\rho,\pi,S',\Nop}}
\quad
\inferrule[I-Cat\label{rule:i-cat}]
{S = s_2::s_1::S'}
{\Angle{\rho,\pi,S,\Cat}\xrightarrow{\epsilon}\Angle{\rho,\pi,(s_1\cdot s_2)::S',\Nop}}
\\
\inferrule[I-Template\label{rule:i-template}]
{s = t[\Brace{x}\mapsto \rho(x)\mid x\in\Dom{\rho}]
    [\Brace{x}\mapsto \sigma(x)\mid x\in\Dom{\sigma}\setminus\Dom{\rho}]}
{\Angle{\rho,\pi,S,\TemplateIR{t}}\xrightarrow{\epsilon}\Angle{\rho,\pi,s::S,\Nop}}
\quad
\inferrule[I-And\label{rule:i-and}]
{S = s_2::s_1::S'\quad s = s_1~\Text{and}~s_2}
{\Angle{\rho,\pi,S,\AndOp}\xrightarrow{\epsilon}\Angle{\rho,\pi,s::S',\Nop}}
\\
\inferrule[I-Or\label{rule:i-or}]
{S = s_2::s_1::S'\quad s = s_1~\Text{or}~s_2}
{\Angle{\rho,\pi,S,\OrOp}\xrightarrow{\epsilon}\Angle{\rho,\pi,s::S',\Nop}}
\quad
\inferrule[I-Not\label{rule:i-not}]
{S = s::S' \quad s' = \Text{not}~s}
{\Angle{\rho,\pi,S,\NotOp}\xrightarrow{\epsilon}\Angle{\rho,\pi,s'::S',\Nop}}
\quad
\inferrule[I-Seq\label{rule:i-seq}]
{\Angle{\rho,\pi,S,ir_1}\xrightarrow{\ell}\Angle{\rho', \pi', S', ir_1'}}
{\Angle{\rho, \pi, S, \Seq{ir_1}{ir_2}}\xrightarrow{\ell}\Angle{\rho', \pi', S',\Seq{ir_1'}{ir_2}}}
\\
\inferrule[I-Nop\label{rule:i-nop}]
{~}
{\Angle{\rho,\pi,S,\Seq\Nop{ir}}\xrightarrow{\epsilon}\Angle{\rho, \pi, S, ir}}
\quad
\inferrule[I-Branch-T\label{rule:i-branch-t}]
{S = \Text{true}::S'}
{\Angle{\rho,\pi,S,\Branch{ir_1}{ir_2}}\xrightarrow{\epsilon}\Angle{\rho,\pi,S',ir_1}}
\quad
\inferrule[I-Branch-F\label{rule:i-branch-f}]
{S = \Text{false}::S'}
{\Angle{\rho,\pi,S,\Branch{ir_1}{ir_2}}\xrightarrow{\epsilon}\Angle{\rho,\pi,S',ir_2}}
\\
\inferrule[I-While\label{rule:i-while}]
{~}
{\Angle{\rho, \pi, S, \IRWhile{ir_1}{ir_2}}\xrightarrow{\epsilon}\Angle{\rho, \pi, S, \Seq{ir_1}{\Branch{\Seq{ir_2}{\IRWhile{ir_1}{ir_2}}}{\Nop}}}}
\end{mathparpagebreakable}
\captionof{figure}{Operational semantics for executing the stepwise IR.}
\label{fig:stepwise-sem}
}

\begin{figure}[htbp!]
    \centering
    \includegraphics[width=.9\textwidth]{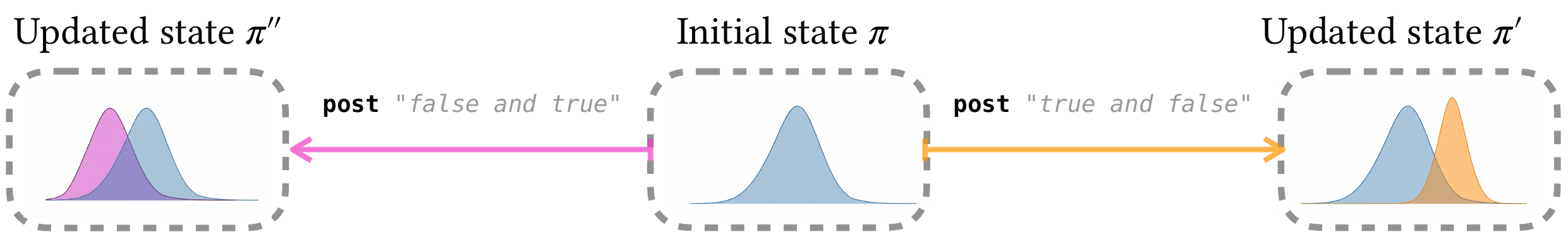}   
    \caption{Evaluation order matters}\label{fig:stepwise-order}
\end{figure}

\paragraph{Prompting effects}
In the classic imperative language $\IMP$, 
expression evaluation is \emph{pure},
as it returns a value without changing $\Store$.
In our $\DSL$, expression evaluation is \emph{effectful},
in the sense that $\ReasonExpr{e}$ and $\CastExpr{e}{\tau}$ invoke $\Post$,
and \Rule{rule:i-post}{I-Post} is the only rule that modifies the prompt history.
\Cref{fig:stepwise-order} gives an example of this effect.
With value stacks $S_1=\Text{true}::\Text{false}::S_0$ and $S_2=\Text{false}::\Text{true}::S_0$,
executing $\AndOp$ yields different execution paths that change the distribution in different ways and may not merge back to the same state.

We call this kind of effect \emph{prompting effects}.
Aside from prompting effects, expression evaluation itself does not involve changes to variable state, which is consistent with $\IMP$.
$\DSLIR$ makes this explicit: 
we compile expressions to $\DSLIR$ sequences, so that pure parts become $\Push{s}$, $\Lookup{x}$, $\TemplateIR{t}$, and $\Cat$, while effectful parts become $\Post$, $\Gen$, or $\Cgen{\tau}$.

\Rule{rule:i-gen}{I-Gen} and \Rule{rule:i-cgen}{I-Cgen} are the rules that involve decoding the LM response.
Given a fixed decoding function, 
$\Decode{\D[\pi]}{\varphi}$ performs constrained decoding under distribution $\D[\pi]$ induced by prompt history $\pi$ and string constraint $\varphi$, 
and returns a string $s$ that is then pushed onto the stack.
As $\Gen$ does not specify any constraint, 
$\Decode{\D[\pi]}{\top}$ simply returns an unconstrained sample from $\D[\pi]$.
It is worth noting that \Rule{rule:i-print}{I-Print} shows that under the current compilation, not all LM responses are presented to the user.

\begin{definition}
\label{def:stepwise}
With all the details set up, 
we can define the stepwise compilation $\SComp$ to be a function that takes $\DSL$ program $P$ and a global environment $\sigma$ and returns the initial state $q_0$ of the transition system
\vspace{-1ex}
\[
\SComp(P,\sigma) = \Angle{\emptyset,\epsilon,[],\Sem{P}\sigma}
\]

Indeed, with this IR, 
the baseline compilation can be viewed abstractly as collapsing the entire program into a single LM step under the global constraint:
\[
\BComp(P,\sigma) \simeq \Seq{\Push{\D\Brack{\Sem{P}\sigma}}}{\Seq{\Post}{\Cgen{\C[P]}}}
\]
where $\Sem{P}$ is the transformation function defined in~\Cref{sec:baseline:1},
and $\C[P]$ is the global string constraint defined in~\Cref{sec:baseline:2}.
\end{definition}

\paragraph{Extending the IR with context management}
In the semantics above, the context $\pi$ grows monotonically:
\Rule{rule:i-post}{I-Post} only ever appends to it, so every LM call observes an increasing prompt history.
This is a deliberate minimality choice that keeps the design space small, but it means strategies requiring \emph{isolated} or \emph{reset} contexts, such as best-of-$N$ sampling or verification in a fresh context, cannot be expressed directly.
The IR extends naturally to accommodate them by adding context-management primitives, for example a reset (or save/restore) operator that checkpoints and restores $\pi$, or a context-subtraction operator that removes a previously posted segment.
We keep such operators out of the core to avoid an explosion of design choices and leave a systematic treatment of context algebras to future work.

\section{Implementation and Evaluation}\label{sec:eval}

We seek to answer the following research questions:
\begin{enumerate}[label=\textbf{RQ\arabic*)}, leftmargin=*]
    \item How do different compilation strategies affect the performance of $\DSL$ programs?
    \item What is the value of typing in $\DSL$ programs?
    \item What are the benefits of using $\DSL$ programming compared to directly using natural language?
    \item What are the limitations of $\DSL$ in terms of expressiveness and applicability?
\end{enumerate}

We begin by describing the implementation of $\DSLPy$ in~\Cref{sec:eval:1} and the shared experimental setup in~\Cref{sec:eval:2}. 
We then present the two case studies, statutory reasoning and data labeling/transformation, 
in~\Cref{sec:eval:2:1,sec:eval:2:2}, emphasizing benchmark-specific workflow details and high-level empirical observations. The answers to RQ1--RQ4 are then distilled in~\Cref{sec:eval:3}, which summarizes the main takeaways from both case studies.

\subsection{Implementation}\label{sec:eval:1}

We implement $\DSL$ as an embedded DSL in Python,
called $\DSLPy$, and use it to write a variety of prompt-based programs that interact with LMs.
User programs in $\DSLPy$ are written as ordinary Python functions annotated with the \texttt{@dsl} decorator.
Since Python uses eager evaluation, 
our implementation rewrites user programs at the Python AST level to construct $\DSL$'s internal syntax tree.
\Cref{fig:impl-workflow} shows the overall workflow of the implementation.

\begin{figure}[htbp!]
\centering
\scalebox{.52}{
\begin{tikzpicture}[
    node distance=0.55cm and 0.55cm,
    box/.style={
        rounded corners=5pt,
        dashed,
        line width=0.9pt,
        draw=black!35,
        fill=white,
        align=center,
        minimum height=1.0cm,
        minimum width=2.6cm,
        font=\Large,
        inner sep=6pt
    },
    stage/.style={box, text width=2.45cm},
    analysisbox/.style={box, text width=2.85cm},
    outbox/.style={box, text width=2.55cm},
    smallbox/.style={
        fill=black!5,
        align=center,
        minimum height=0.55cm,
        minimum width=3.2cm,
        font=\Large,
        inner sep=4pt
    },
    groupbox/.style={
        rounded corners=5pt,
        dashed,
        line width=1.0pt,
        draw=black!40,
        fill=white,
        inner sep=10pt
    },
    note/.style={
        fill=black!5,
        align=center,
        text width=3.25cm,
        font=\large,
        inner sep=4pt
    },
    arrow/.style={
        -{Stealth[length=3mm,width=1.6mm]},
        line width=1.0pt,
        draw=black!55,
        shorten <=1pt,
        shorten >=1pt
    }
]

\node[stage] (src) {Python source\\with \texttt{@dsl}};
\node[stage, right=of src] (rewrite) {AST rewrite};
\node[stage, right=of rewrite] (build) {Internal $\DSL$\\AST construction};
\node[stage, right=of build] (norm) {Normalization};
\node[analysisbox, right=of norm] (analysis) {Type \& constraint\\analysis};
\node[groupbox, right=0.75cm of analysis] (runnergroup) {
    \begin{tikzpicture}[node distance=0.18cm]
        \node[font=\normalsize\bfseries, draw=none, fill=none] (title) {Compiler + executor};
        \node[smallbox, below=0.2cm of title] (baseline) {Baseline};
        \node[smallbox, below=0.18cm of baseline] (stepwise) {Stepwise};
    \end{tikzpicture}
};
\node[outbox, right=0.75cm of runnergroup] (out) {Outputs / return};
\draw[arrow] (src) -- (rewrite);
\draw[arrow] (rewrite) -- (build);
\draw[arrow] (build) -- (norm);
\draw[arrow] (norm) -- (analysis);
\draw[arrow] (analysis) -- (runnergroup);
\draw[arrow] (runnergroup) -- (out);
\node[note, above=0.18cm of rewrite] {
Python syntax $\to$ explicit $\DSL$ constructors};
\node[note, below=0.18cm of norm] {
\texttt{get}/\texttt{infer} $\to$ \texttt{GetStmt}
};
\end{tikzpicture}
}
\vspace{3ex}
\caption{Implementation workflow of $\DSLPy$.}
\label{fig:impl-workflow}
\vspace{-2em}
\end{figure}

\begin{figure}[htbp!]
\centering
\begin{minipage}[t]{.62\textwidth}
    \vspace{1em}
    \centering
    \includegraphics[width=\textwidth]{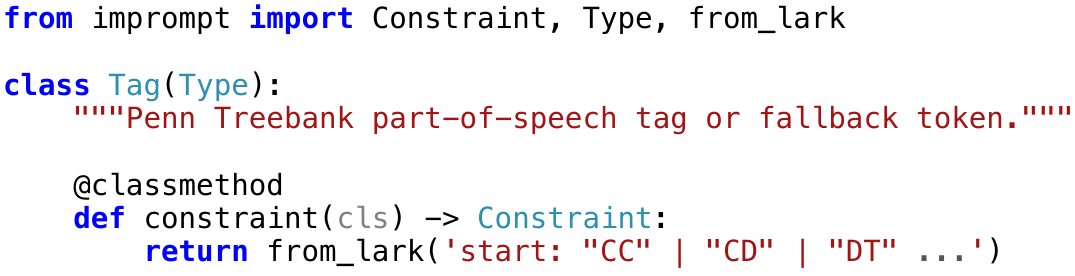}
    \caption{Types via Python classes in $\DSLPy$}\label{fig:tag-dslpy}
\end{minipage}
\begin{minipage}[t]{.36\textwidth}
    \vspace{0pt}
    \centering
    \tiny
    \begin{algorithmic}[1]
    \AlgoFunction{\textsc{ConstrainedDecode}}{$U, G, T, p$}
        \AlgoState $M \gets \textsc{Matcher}(G)$; $\t \gets \epsilon$
        \AlgoWhile{$\neg M.\textsc{Accepting}()$}
        \AlgoState $\ell \gets \textsc{ModelStep}(U,\t)$
        \AlgoState $\ell \gets \textsc{Mask}(\ell, M)$
        \AlgoState $t \gets \textsc{Sample}(\ell; T, p)$
        \AlgoState $\t \gets \t \cdot t$; $M.\textsc{Consume}(t)$
        \AlgoEndWhile
        \AlgoState\AlgoReturn $\textsc{Decode}(\t)$
    \AlgoEndFunction
    \end{algorithmic}
    \caption{Constrained decoding}
    \label{fig:impl-baseline-decoding}
\end{minipage}
\end{figure}

\paragraph{Types and constraints.}
As mentioned in~\Cref{sec:lang}, 
user-defined types $\Ty{\omega,\varphi}$ are implemented as a Python class extending a base class \texttt{Type}, 
with the textual description $\omega$ given by the class docstring and the constraint $\varphi$ given by a method \texttt{constraint}.
Grammar constraints returned by the method \texttt{constraint} are described in \verb|Lark|~\cite{lark} and enforced during decoding by \verb|llguidance|~\cite{guidance2025}.
In baseline mode, the implementation combines the relevant type constraints into a single response grammar, as described abstractly in~\Cref{sec:baseline:2}.
For example,~\Cref{fig:tag-dslpy} shows a user-defined type \texttt{Tag} in our running PoS example. Note that the ``$\Text{...}$'' is for presentation simplicity, 
not part of the \verb|Lark| syntax.

\Cref{fig:impl-baseline-decoding} shows the core constrained-decoding loop.
$\textsc{Matcher}$ operationalizes the decidability of the constraint $\varphi$.
Each step samples only from tokens permitted by that matcher state, 
using the configured temperature $T$ and top-$p$ parameters, 
and the implementation enforces a decoding timeout.

\begin{table}[htbp!]
    \centering
    \fontsize{8}{9}\selectfont
    \begin{tabular}{p{0.18\textwidth} p{0.20\textwidth} p{0.53\textwidth}}
        \toprule
        $\DSL$ syntax & $\DSLPy$ syntax & Internal treatment \\
        \midrule
        $\ReasonExpr{e}$ & \texttt{reason(expr)} & Constructs a \texttt{ReasonExpr} with dynamic result type. \\
        $\CastExpr{e}{\tau}$ & \texttt{expr @ T()} & Lowers to a typed reinterpretation with constrained decoding. \\
        $\LetEqStmt{x}{e}$ & \texttt{x = expr} & Rewritten to \texttt{LetStmt(Var("x"), expr)}. \\
        $\SayStmt{e}$ & \texttt{say(expr)} & Emits visible program output. \\
        $\IfStmt{b}{c_1}~\ElseStmt{c_2}$ & \texttt{if expr: ... else:} & Rewritten into explicit block statements. \\
        $\WhileStmt{b}{c}$ & \texttt{while expr: ...} & Rewritten into an explicit \texttt{WhileStmt}. \\
        $\GetStmt{x\Colon\tau}{t}$ & \texttt{x = get(T()); infer(f"...\{x\}...")} & Normalized into a first-class \texttt{GetStmt}. \\
        \bottomrule
    \end{tabular}
    \caption{Correspondence between $\DSL$ and $\DSLPy$ statements}\label{tab:impl-syntax-map}
\end{table}

\vspace{-1em}
\paragraph{Frontend and core syntax.}
We show the correspondence between the abstract syntax of $\DSL$ and its Python embedding in~\Cref{tab:impl-syntax-map}.
The frontend rewrites Python syntax into explicit $\DSL$ constructors. It normalizes the surface pattern \texttt{x = get(T())} together with \texttt{infer("...\{x\}...")} into a first-class \texttt{GetStmt}, and it also supports first-class function definitions and calls via \texttt{@dsl}. 
We show this syntax concretely in~\Cref{sec:eval:2:1}.

\subsection{Evaluation Setup}\label{sec:eval:2}

\paragraph{Benchmarks}
We evaluate $\DSL$ on two case studies: 
statutory reasoning and data labeling/transformation. 
The statutory-reasoning case study uses 114 distinct statute-level functions over 9 statute files (sections), 
exercised on 276 cases. 
The data-labeling case study uses 5 PromptPex programs with 67 tests. 
We describe the two benchmarks and their task structure in more detail in~\Cref{sec:eval:2:1,sec:eval:2:2}.

\paragraph{Experiment setup}
We run all experiments on a remote server with PyTorch 2.8.0 and Python 3.12 (Ubuntu 22.04) with CUDA 12.8, equipped with a single 48GB vGPU, 20 vCPUs (Intel Xeon Platinum 8470Q), and 96GB RAM.
We use Hugging Face Transformers~\cite{wolf-etal-2020-transformers} for local model inference, 
and \verb|llguidance| for token mask computation and constraint decoding.

We evaluate four execution modes over the same source programs.
\emph{Baseline} compiles the entry function together with all transitively reachable functions into a prompt template with output
constraints.
\emph{Baseline-WC} uses the same compiled prompt, but removes the output constraints.
\emph{Stepwise} executes the program through $\DSLIR$ as a sequence of smaller constrained calls, with standard handling for function
arguments and function calls.
\emph{Natural} bypasses $\DSL$ compilation and prompts the model directly using the task description provided by the benchmark. 
For the first case study,
this prompt consists of the relevant statutory text together with the case facts and benchmark question. 
For the second case study, 
we use the original natural-language task description supplied by the dataset. 
Details are described in~\Cref{sec:eval:2:1,sec:eval:2:2}.

All experiments are run on three models: Llama3.2-1B-Instruct, Phi3.5-Mini-Instruct, and Gemma2-9B-IT. We use temperature $0.7$, top-$p$
$0.9$, and a timeout of 120 seconds for each LM call. For each test case, we run the system three times and report the average.

\subsection{Case Study 1: Statutory Reasoning}\label{sec:eval:2:1}

We evaluate $\DSL$ on SARA~\cite{DBLP:conf/kdd/HolzenbergerBD20}, 
a benchmark for statutory reasoning over selected provisions of the U.S.
Internal Revenue Code (IRC). 
Our subset covers 276 cases drawn from nine IRC sections: 
tax liability (\S1, \S3301), 
filing status (\S2, \S7703),
dependents (\S152), 
and deductions and exemptions (\S63, \S68, \S151, \S3306). 
Each case provides a factual scenario, 
a natural-language yes/no question, 
and a ground-truth Boolean answer.

\begin{figure}[htbp!]
    \centering
    \begin{subfigure}{0.45\textwidth}
    \centering
    \vspace{2em}
    \scalebox{0.7}{
    
  \begin{tikzpicture}[
      node distance=1.9cm and 1.5cm,
      every node/.style={
          rounded corners=3pt,
          minimum width=1.35cm,
          minimum height=0.62cm,
          align=center,
          draw=none,
          fill=myMainRed!18
      },
      dep/.style={-{Stealth[length=2.6mm,width=1.8mm]}, line width=1.3pt, draw=myArrowRed}
      ]
      \node (s1) {\S1};
      \node[right=of s1] (s63) {\S63};
      \node[right=of s63=.8cm] (s3301) {\S3301};

      \node[below=of s1] (s2) {\S2};
      \node[below=of s63] (s151) {\S151};
      \node[below=of s3301] (s3306) {\S3306};

      \node[below left=1.1cm and 0.8cm of s151] (s7703) {\S7703};
      \node[below=of s151] (s152) {\S152};
      \node[below right=1.1cm and 0.8cm of s151] (s68) {\S68};

      \draw[dep] (s1) -- (s2);
      \draw[dep] (s1) to[out=230, in=150] (s7703);
      \draw[dep] (s63) -- (s2);
      \draw[dep] (s63) -- (s151);
      \draw[dep] (s63) -- (s68);
      \draw[dep] (s151) -- (s152);
      \draw[dep] (s68) -- (s2);
      \draw[dep] (s68) -- (s7703);
      \draw[dep] (s2) -- (s7703);
      \draw[dep] (s3301) -- (s3306);
    \end{tikzpicture}
    }
    \caption{Section-level dependency graph}
    \label{fig:sara-deps:1}
    \end{subfigure}
    \begin{subfigure}{0.45\textwidth}
    \includegraphics[width=\linewidth]{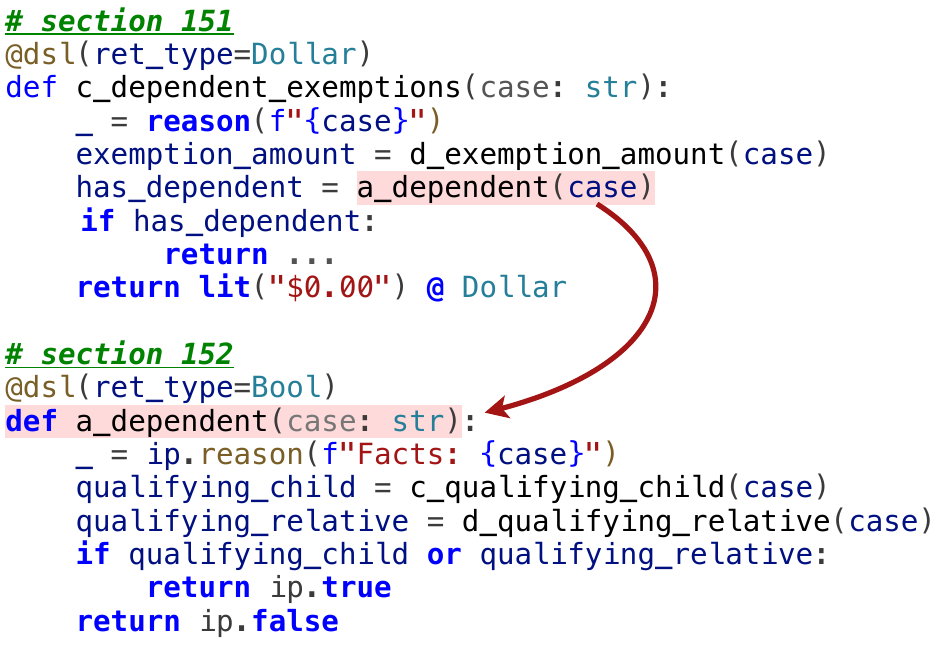}
    \caption{Function-level dependency example}
    \label{fig:sara-deps:2}
    \end{subfigure}
    \caption{Dependencies in SARA benchmark}\label{fig:sara-deps}
\end{figure}

SARA is a challenging task because the benchmark questions cannot be answered using simple lookups. 
The answers require complex computations that are performed over a mutually dependent set of statutory predicates present across IRC sections.
\Cref{fig:sara-deps:1} shows the section-level structure in our $\DSLPy$ encoding.
\Cref{fig:sara-deps:2} shows a concrete cross-section dependency:
\S151(c) grants an additional exemption only for an individual who is ``a dependent (as defined in section 152)'' of the taxpayer. 
Accordingly, in our $\DSLPy$ encoding, 
\texttt{c\textbf{\_}dependent\textbf{\_}exemptions} in \S151 calls \texttt{a\textbf{\_}dependent} in \S152.

\vspace{-1em}
\paragraph{Workflow.}
\Cref{fig:sara-workflow} summarizes the evaluation pipeline. 
Each case file is mapped to one statute function. 
In \emph{Baseline}, \emph{Baseline-WC}, and \emph{Stepwise} modes, 
we execute the corresponding $\DSLPy$ program; 
in natural mode, we instead ask the model directly using the statutory text and the original benchmark question. 

\begin{figure}[htbp!]
    \centering
    \scalebox{.69}{
        \begin{tikzpicture}[
          node distance=0.5cm and 0.5cm,
          box/.style={
              rounded corners=5pt,
              dashed,
              line width=0.9pt,
              draw=black!35,
              fill=white,
              align=center,
              minimum height=1.0cm,
              minimum width=2.7cm,
              font=\Large,
              inner sep=6pt
          },
          smallbox/.style={
              fill=black!5,
              align=center,
              minimum height=0.55cm,
              minimum width=2.6cm,
              font=\Large,
              inner sep=4pt,
          },
          groupbox/.style={
              rounded corners=5pt,
              dashed,
              line width=1.0pt,
              draw=black!40,
              fill=white,
              inner sep=10pt
          },
          arrow/.style={
              -{Stealth[length=3mm,width=1.6mm]},
              line width=1.0pt,
              draw=black!55,
              shorten <=1pt,
              shorten >=1pt
          }
      ]
      \node[box] (case) {SARA case\\facts + question};
      \node[groupbox, right=of case] (runnergroup) {
        \begin{tikzpicture}[node distance=0.18cm]
            \node[font=\normalsize\bfseries, draw=none, fill=none] (title) {Statute runner};
            \node[smallbox, below=0.2cm of title] (natural) {\emph{Natural}};
            \node[smallbox, below=0.18cm of natural] (baseline) {\emph{Baseline}};
            \node[smallbox, below=0.18cm of baseline] (baselinewc) {\emph{Baseline-WC}};
            \node[smallbox, below=0.18cm of baselinewc] (stepwise) {\emph{Stepwise}};
        \end{tikzpicture}
    };
    \node[box, right=of runnergroup, minimum width=3.6cm] (judge) {Adjudication layer\\function result $\to$ benchmark answer};
    \node[box, right=of judge, minimum width=2.9cm] (ans) {Final label\\\Text{true} / \Text{false} / \Text{unanswered}};
    \draw[arrow](case) -- (runnergroup);
    \draw[arrow](runnergroup) -- (judge);
    \draw[arrow](judge) -- (ans);
    \end{tikzpicture}
    }
    \caption{SARA evaluation workflow.}
    \label{fig:sara-workflow}
    \vspace{-1em}
\end{figure}

The last step of converting the result into a boolean is not a trivial one, %
because the benchmark question and the function output are often not at the same level of abstraction. 
For example, a function may compute an amount such as the standard deduction or the applicable amount under \S68, while the related benchmark question asks whether a particular statutory proposition is true. 
Across the 276 cases, this yields four groups: 
166 direct boolean questions, 
84 amount-comparison questions, 
4 exemption-amount-to-boolean questions, 
and 22 manual bridge-rule questions. 
Only the direct boolean cases are deterministic from the raw function return; 
for the other questions,
we issue one additional request to the same model, 
asking it to convert the function result into the final boolean benchmark answer (or \verb|unanswered|).
In SARA, we do not use a single external expert model for this adjudication step.
because our goal is to evaluate end-to-end task performance in a model-specific way, and introducing a separate judge model would confound that measurement.
Using the same model for the final conversion avoids adding an additional source of cross-model bias.

We report on three metrics:
\[
\textit{success}=\frac{\#\text{ok}}{\#\text{total}},\qquad
\textit{coverage}_t=\frac{\#\text{answered}}{\#\text{ok}},\qquad
\textit{accuracy}=\frac{\#\text{correct}}{\#\text{total}}.
\]
Here, $\#\text{total}$ is the total number of test cases. $\#\text{ok}$ counts runs that complete successfully and return a result,
$\#\text{answered}$ counts runs whose returned result is judged relevant to the question,
and $\#\text{correct}$ counts runs that return the correct answer.
Intuitively, \textit{success} filters out failures such as timeouts due to an excessively large decoding search space or other execution errors.
\textit{coverage}$_t$ further excludes cases where the run succeeds but the response is completely irrelevant (e.g., repeating the question without answering it).
\textit{accuracy} measures the overall correctness on all test cases.

\begin{figure}[htbp!]
    \centering
    \makebox[\textwidth][c]{%
        \includegraphics[width=.5\textwidth]{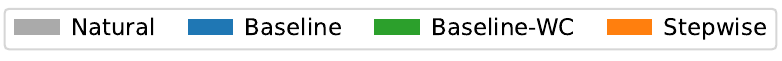}%
    }
    \begin{subfigure}{0.49\textwidth}
        \centering
        \raisebox{.2em}{%
        \includegraphics[width=\linewidth]{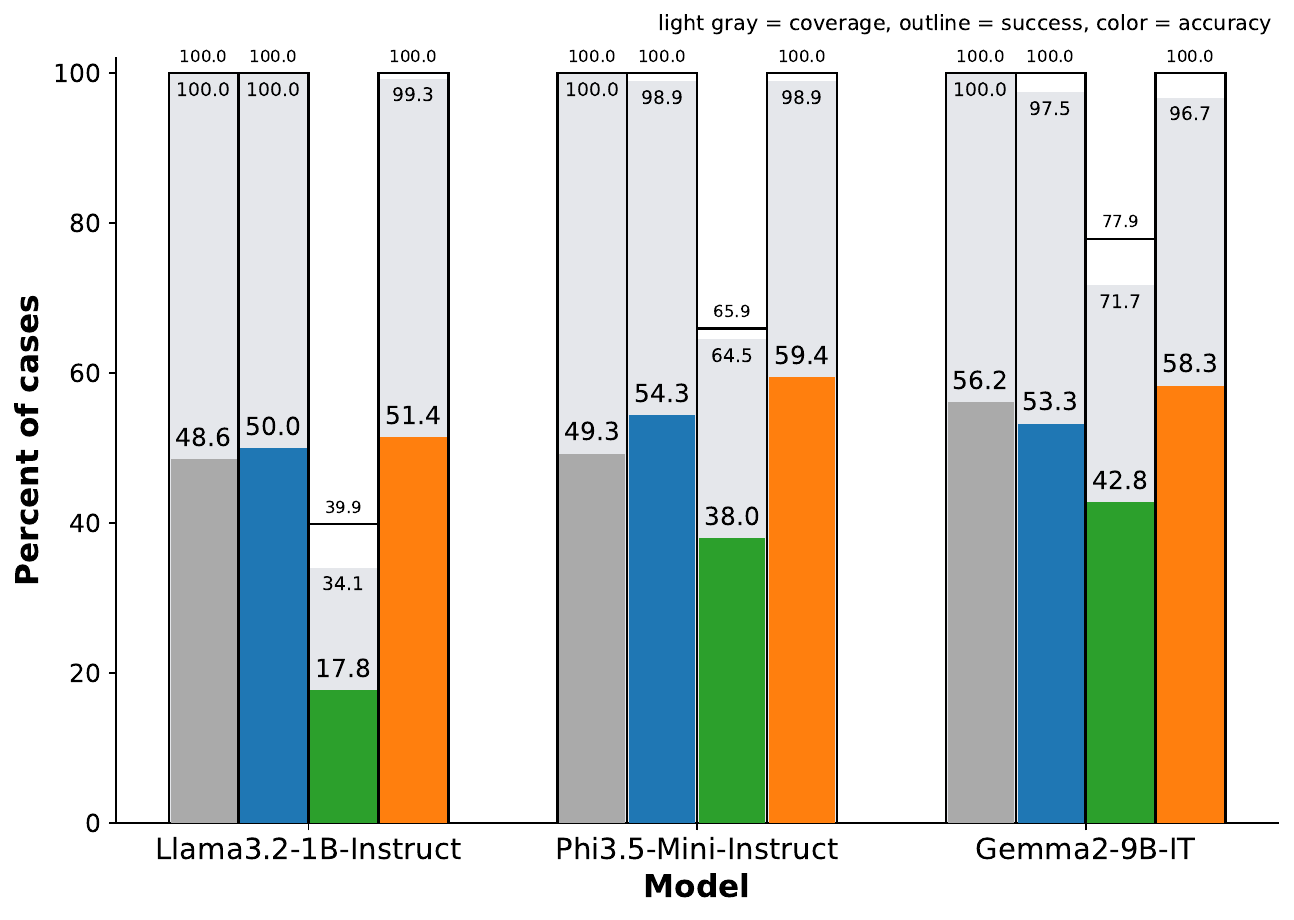}
        }
        \caption{Accuracy by compilation strategy.}\label{fig:sara-modes}
    \end{subfigure}
    \begin{subfigure}{0.49\textwidth}
        \centering
        \includegraphics[width=\linewidth]{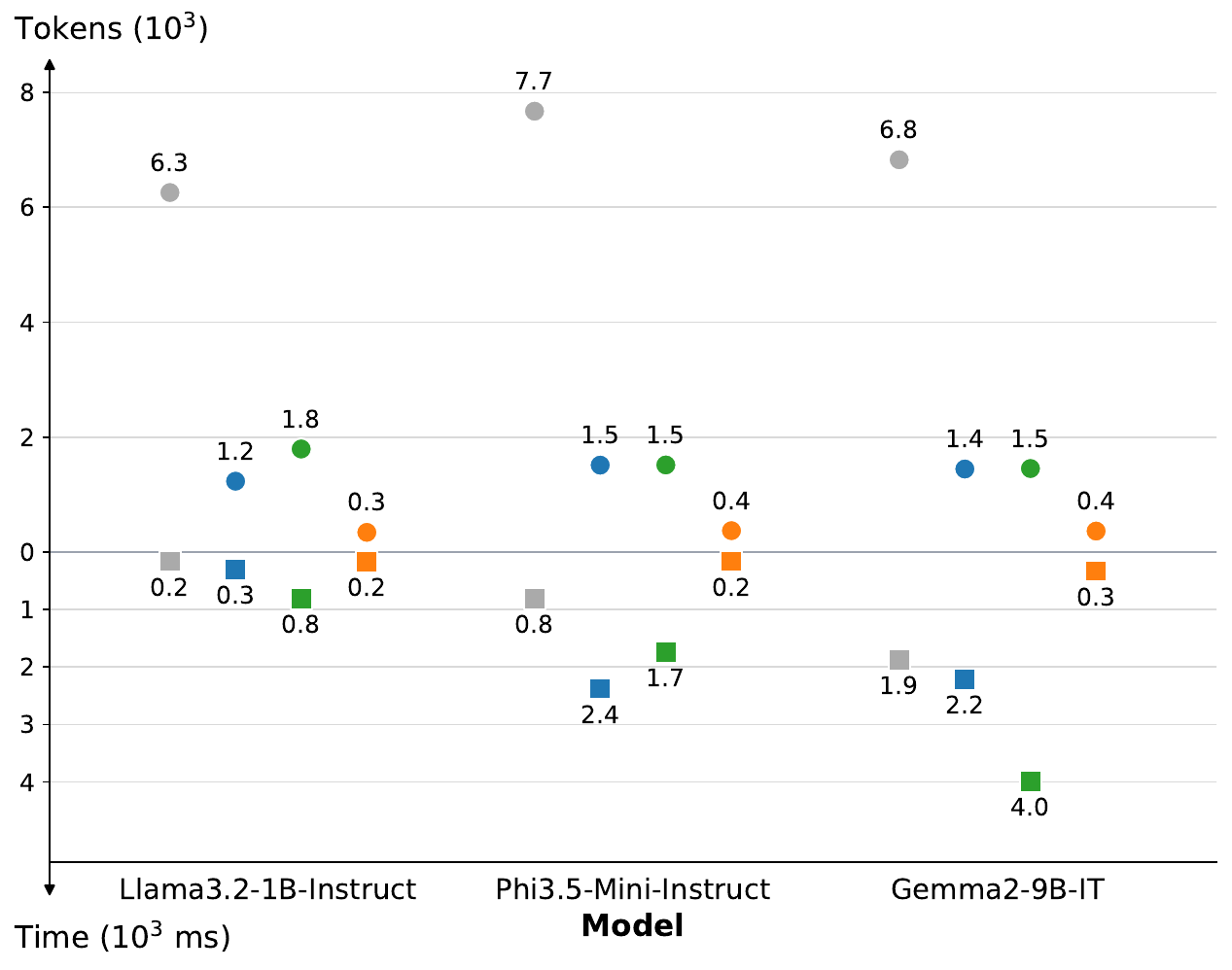}
        \caption{Median token and latency cost on successful cases.}\label{fig:sara-efficiency}
    \end{subfigure}
    \caption{SARA results.}\label{fig:sara-results}
\end{figure}

\paragraph{Results.}
\Cref{fig:sara-results} summarizes the SARA results: \Cref{fig:sara-modes} compares end-to-end accuracy across execution modes, 
and \Cref{fig:sara-efficiency} reports median latency and token usage on successful runs. Across all three models, \emph{stepwise} achieves the best end-to-end accuracy: 51.4\% vs.~50.0\% (\emph{baseline}) and 48.6\% (\emph{natural}) on Llama 3.2 1B, 59.4\% vs.~54.3\% and 49.3\% on Phi 3.5 Mini, 
and 58.3\% vs.~53.3\% and 56.2\% on Gemma 2 9B. 
Aggregated over all 276 cases by majority vote across the three models, \emph{stepwise} remains ahead at 59.1\% (163/276), versus 54.3\% (150/276) for \emph{baseline} and 52.9\% (146/276) for \emph{natural}.
Both \emph{Baseline} and \emph{Stepwise} maintain perfect function-level success and high coverage, 
while \emph{Baseline-WC} degrades sharply, indicating that type-directed constraints are doing substantial work.
\emph{Stepwise} is also cheaper, 
using about 343--373 median tokens versus roughly 1.2k--1.5k for \emph{Baseline}, 
with median latency of 162--329ms versus 307ms--2.38s.

\subsection{Case Study 2: Data Labeling and Transformation}\label{sec:eval:2:2}
Our examples in this section are drawn from PromptPex~\cite{promptpexsharma2025automatictestgeneration}, an automated tool for generating prompt test cases.
We select five tasks with programmatic aspects:
\One~text classification (\texttt{classify-input-text}), 
\Two~structured extraction (\texttt{elements}, 
\Three~\texttt{extract-names}), 
\Four~PoS tagging (\texttt{speech-tag}), 
and \Five~HTML-style text transformation (\verb|text-to-p|).

\begin{figure}[htbp!]
    \centering
    \vspace{-.5em}
    \includegraphics[width=0.9\textwidth]{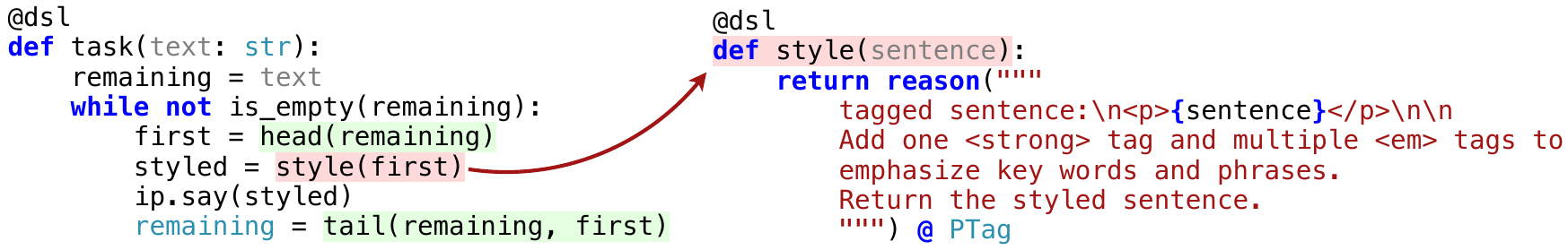}
    \caption{Example: \texttt{text-to-p} core function}\label{fig:while}
    \vspace{-1em}
\end{figure}

\begin{figure}[htbp!]
    \centering
    \includegraphics[width=.9\textwidth]{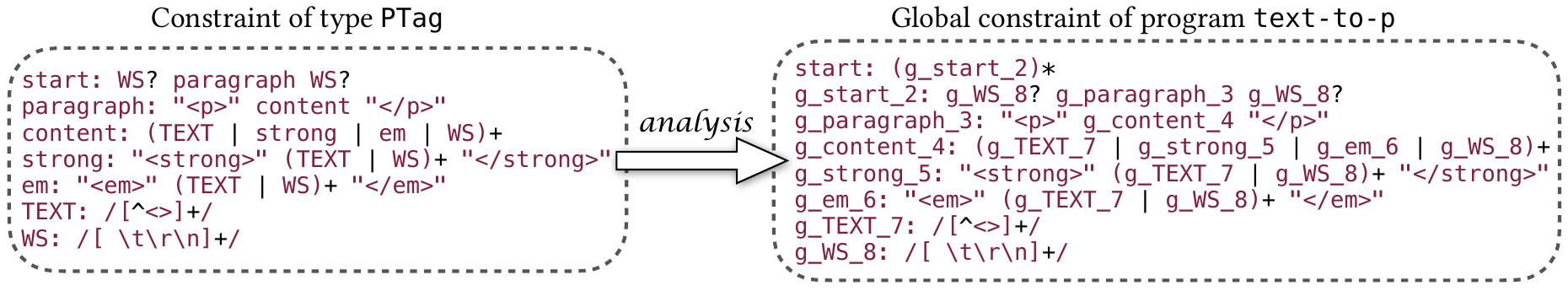}
    \caption{Type \texttt{PTag} in \texttt{text-to-p} problem}\label{fig:ptag}
    \vspace{-1em}
\end{figure}

Compared to~\Cref{sec:eval:2:1}, 
tasks in this case study are shorter and their outputs are much more tightly constrained.
In SARA, the main challenge is long-range statutory reasoning over larger prompts.
In PromptPex, the challenge is often to satisfy strong local string constraints.
For example, the short program in~\Cref{fig:while} asks the LM to split a paragraph into individual sentences and wrap each sentence with a \texttt{<p>} tag,
and \emph{each} paragraph must contain both \texttt{<strong>} and multiple \texttt{<em>} tags.
Here, the requirement \emph{each paragraph} is implemented via a $\DSLPy$ $\While$ loop,
and the per-sentence formatting is expressed by calling the $\DSLPy$ \sethlcolor{myRed}\hl{\texttt{style}} function.
Notably, to ensure a fair comparison, \sethlcolor{myGreen}\hl{\texttt{head}} and \sethlcolor{myGreen}\hl{\texttt{tail}} are also $\DSLPy$ functions rather than Python string operations.
\Cref{fig:ptag} shows the string constraint $\C[\texttt{PTag}]$ expressed in \verb|Lark| syntax, 
together with the global string constraint generated by our analysis.

\paragraph{Workflow and metrics}
Each PromptPex test provides candidate inputs and expected outputs, 
but the correctness of a model response is still not deterministic.
For example, in the \texttt{text-to-p} task, the LM can add tags in multiple valid ways that satisfy the requirements.
In our evaluation, we use GPT-5.4 as a judge model: given the test input/output and the task description, it determines whether the output meets the specification.
Since these tasks are short, \textit{success} and \textit{coverage} are both 100\%, so we report only \textit{accuracy}.
For each test, we run three trials and report the mean accuracy.

\begin{figure}[htbp!]
    \centering
    \makebox[\textwidth][c]{%
        \includegraphics[width=.5\textwidth]{figures/legend.pdf}%
    }
    \begin{subfigure}{0.49\textwidth}
        \centering
        \raisebox{.2em}{%
        \includegraphics[width=\linewidth]{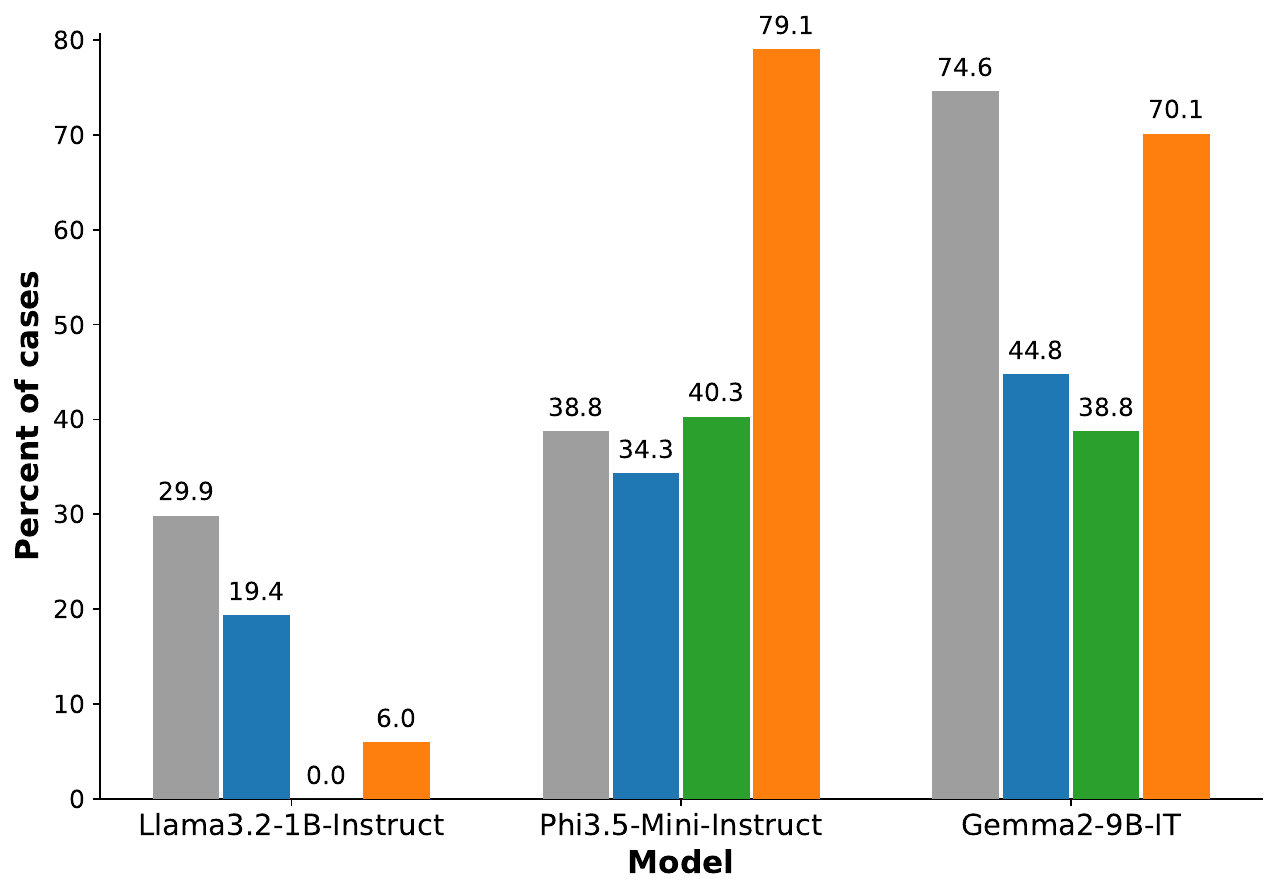}
        }
        \caption{Accuracy by compilation strategy.}\label{fig:promptpex-modes}
    \end{subfigure}
    \begin{subfigure}{0.49\textwidth}
        \centering
        \includegraphics[width=\linewidth]{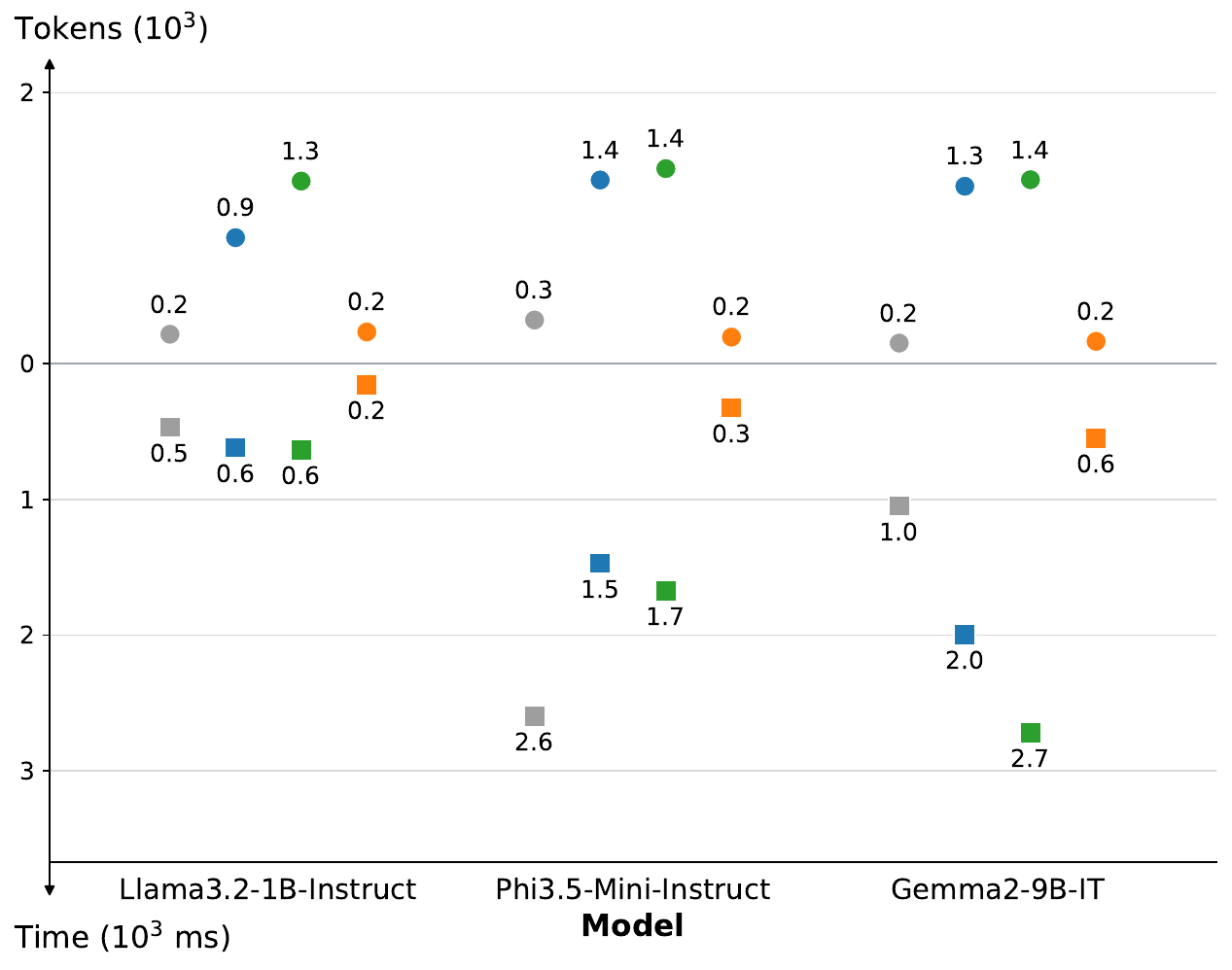}
        \caption{Median token and latency cost on successful cases.}\label{fig:promptpex-efficiency}
    \end{subfigure}
    \caption{PromptPex overall results.}\label{fig:promptpex-results}
\end{figure}

\begin{table*}[htbp!]
    \centering
    \scalebox{.9}{
    \fontsize{8}{8}\selectfont
    \setlength{\tabcolsep}{3pt}
      \begin{tabular}{p{3.7cm}|*{12}{>{\centering\arraybackslash}p{0.62cm}}}
        \toprule
        \multicolumn{1}{c}{Task} & \multicolumn{4}{c}{Llama3.2-1B-Instruct} & \multicolumn{4}{c}{Phi3.5-Mini-Instruct} & \multicolumn{4}{c}{Gemma2-9B-IT} \\
        & \multicolumn{1}{c}{N}
        & \multicolumn{1}{c}{B}
        & \multicolumn{1}{>{\centering\arraybackslash}p{0.9cm}}{B-WC}
        & \multicolumn{1}{c}{S}
        & \multicolumn{1}{c}{N}
        & \multicolumn{1}{c}{B}
        & \multicolumn{1}{>{\centering\arraybackslash}p{0.9cm}}{B-WC}
        & \multicolumn{1}{c}{S}
        & \multicolumn{1}{c}{N}
        & \multicolumn{1}{c}{B}
        & \multicolumn{1}{>{\centering\arraybackslash}p{0.9cm}}{B-WC}
        & \multicolumn{1}{c}{S} \\
        \midrule
        \texttt{classify-input-text} $(n=6)$ & {\sethlcolor{myRed}\hl{\textbf{66.7}}} & 33.3 & 0.0 & 33.3 & 83.3 & 66.7 & 66.7 &
{\sethlcolor{myRed}\hl{\textbf{100.0}}} & {\sethlcolor{myRed}\hl{\textbf{100.0}}} & {\sethlcolor{myRed}\hl{\textbf{100.0}}} & 33.3 & {\sethlcolor{myRed}\hl{\textbf{100.0}}} \\
        \texttt{elements} $(n=18)$ & 33.3 & {\sethlcolor{myRed}\hl{\textbf{55.6}}} & 0.0 & 0.0 & 38.9 & 55.6 & 55.6 & {\sethlcolor{myRed}\hl{\textbf{72.2}}} & 44.4 &
72.2 & 66.7 & {\sethlcolor{myRed}\hl{\textbf{77.8}}} \\
        \texttt{extract-names} $(n=12)$ & 0.0 & {\sethlcolor{myRed}\hl{\textbf{8.3}}} & 0.0 & {\sethlcolor{myRed}\hl{\textbf{8.3}}} & 33.3 & 50.0 & 16.7 &
{\sethlcolor{myRed}\hl{\textbf{75.0}}} & {\sethlcolor{myRed}\hl{\textbf{83.3}}} & 58.3 & 33.3 & {\sethlcolor{myRed}\hl{\textbf{83.3}}} \\
        \texttt{speech-tag} $(n=13)$ & 0.0 & 0.0 & 0.0 & {\sethlcolor{myRed}\hl{\textbf{7.7}}} & 38.5 & 23.1 & 53.8 & {\sethlcolor{myRed}\hl{\textbf{61.5}}} &
{\sethlcolor{myRed}\hl{\textbf{92.3}}} & 30.8 & 46.2 & {\sethlcolor{myRed}\hl{\textbf{92.3}}} \\
        \texttt{text-to-p} $(n=18)$ & {\sethlcolor{myRed}\hl{\textbf{55.6}}} & 0.0 & 0.0 & 0.0 & 27.8 & 0.0 & 22.2 & {\sethlcolor{myRed}\hl{\textbf{94.4}}} &
{\sethlcolor{myRed}\hl{\textbf{77.8}}} & 0.0 & 11.1 & 27.8 \\
        \bottomrule
    \end{tabular}
    }
    \caption{PromptPex accuracy (\%) by task, model, and mode. N=Natural, B=Baseline, B-WC=Baseline-WC, S=Stepwise. Bolded values indicate the best performance for each model and task.}
    \label{tab:promptpex-results}
\end{table*}

\paragraph{Results.}
\Cref{fig:promptpex-results} summarizes the aggregate PromptPex results, and \Cref{tab:promptpex-results} breaks them down by task.
PromptPex shows a different pattern from SARA. In \emph{Natural} mode, performance largely follows model size: Gemma2-9B-IT achieves the
highest overall accuracy (74.6\%), followed by Phi3.5-Mini-Instruct (38.8\%) and Llama3.2-1B-Instruct (14.9\%). After compilation, however,
\emph{stepwise} becomes the strongest compiled mode, reaching 79.1\% on Phi3.5-Mini-Instruct and 70.1\% on Gemma2-9B-IT, while
\emph{baseline} is often much worse on tightly constrained tasks such as \texttt{text-to-p}. \emph{Stepwise} is also usually cheaper than
\emph{baseline}: for Gemma2-9B-IT and Phi3.5-Mini-Instruct, median token usage drops from about 1.3k--1.6k to 165--196 tokens, and median
latency drops from about 1.0--2.7s to 325--551ms.

\subsection{Summary by Research Question}\label{sec:eval:3}

\paragraph{RQ1: Compilation Strategies}
Among the two compilation strategies studied in this paper, \emph{stepwise} generally achieves higher accuracy and lower cost in our case
studies. However, this trend is not uniform. We also observe cases where \emph{baseline} performs substantially better than \emph{stepwise},
such as \texttt{elements} with Llama3.2-1B-Instruct. Moreover, the performance of \emph{stepwise} is not always stable across models: on the
same \texttt{text-to-p} task, it performs very well on Phi3.5-Mini (94.4\%) but much worse on the larger Gemma2-9B-IT, even though both
models are instruction-tuned.

\paragraph{RQ2: Value of Typing}
In this paper, we design the type system of $\DSL$ to be closely tied to constraints. Across our case studies, \emph{Baseline} is, in most
cases, both more accurate and cheaper than \emph{Baseline-WC}, which uses the same compilation strategy but without constraints. This
suggests that, in $\DSL$, types can improve both accuracy and efficiency through constraint-guided decoding.

\paragraph{RQ3: $\DSL$ vs Natural Language}
Across most tasks in the two case studies above, 
there exists at least one $\DSL$ compilation strategy that outperforms natural-language
prompting in accuracy. Moreover, in SARA, stepwise compilation also uses substantially fewer tokens and less time than natural prompting.
This suggests that programming with $\DSL$ may improve both effectiveness and efficiency over direct natural-language interaction.

\paragraph{RQ4: Limitations}
$\DSL$ is well suited to procedural tasks with explicit program-style control flow and clear output-structure constraints. However, it also
has several limitations.
First, in terms of expressiveness, the current constraint mechanism is purely syntactic.
As a result, semantic requirements, such as generating text in a Shakespearean style~\cite{promptpexsharma2025automatictestgeneration}, 
are difficult to capture in the current type system.
Second, in terms of applicability, compilation for $\DSL$ should ideally be model-aware, 
but the compilation strategies we present are uniform across models.
For example, our baseline strategy follows Anthropic Claude's recommended XML-style prompting practice~\cite{anthropic2024xmlprompting},
whereas the best prompting format for Google's Gemma family may differ~\cite{google_functiongemma_formatting}.
This suggests that future $\DSL$ compilers may need to adapt their prompt representations to the conventions and strengths of the target model.
If model capabilities and prompting conventions eventually converge across families of LMs, such model-specific adaptation may become
unnecessary.

\section{Discussion and Open Questions}\label{sec:discussion}

Designing a prompt programming language requires some understanding of the low-level primitives exposed by LMs. However, beyond a small number of early attempts to reason about LM execution and control~\cite{meyerson2025solvingmillionstepllmtask,meyerson2025position}, 
there is still no widely accepted account of what the right primitive set should be. 
We therefore view $\DSL$ and its compilers as a first step rather than a complete design. 
We detail below some significant questions about prompt programming and compilation that, 
to the best of our knowledge, remain unanswered: 

\noindent
\textbf{\emph{Equivalence of compilation strategies.}}
Our two compilers can behave differently on the same program, which raises the question of in what sense they realize one language rather than two.
We regard a $\DSL$ program as carrying a single intended meaning, with each strategy an approximation that trades accuracy against cost, much as compiling a program at different optimization levels does not yield two different programs.
Making this precise is difficult because LM execution is stochastic, so two executions may be comparable only in their final result.
One avenue is to model an idealized instruction-tuned model that reads the baseline-compiled prompt with multiple passes or limited lookahead over branching decisions, and to relate it to the stepwise system by an inductive argument that both realize the same intermediate states; recent tools for modeling the computation of language models~\cite{butoi2024computational} may supply the needed structure.
We leave a formal account to future work.

\noindent
\textbf{\emph{Memory.}} 
The two strategies make opposite assumptions about memory, and neither is clearly ideal. Baseline relies on the LM to \emph{remember} variable bindings from the prompt, 
which becomes fragile as prompts grow longer and more complex. 
Stepwise stores bindings in an external environment and shows them to the LM only when needed, 
which avoids trusting the LM as memory but may also make the task less coherent. 
How memory should be represented in prompt programs remains open.

\noindent
\textbf{\emph{Assertions and checks.}}
Prompt programs often benefit from explicit assertions, rubrics, and self-checking stages, but $\DSL$ currently has no dedicated construct for expressing them.

\noindent
\textbf{\emph{Structured data representation.}}
Our current design supports string-based types with a single refinement constraint, but it does not support types with multiple programmatically accessible fields, such as a record with separate \texttt{name} and \texttt{age} fields, each carrying its own refinement.
The concrete challenge is to extend type-directed constraint generation to compositional and possibly nested field structures, while keeping the derived global constraint decidable and the field-level casts reliable.

\noindent
\textbf{\emph{Interactive prompting.}}
The current model assumes a fixed program and input, rather than an ongoing dialogue in which later steps depend on user intervention or other external feedback.

\noindent
\textbf{\emph{Type casting.}}
Our cast operation is intentionally permissive: it delegates the actual mapping from a source type to a target type to the LM, returning a null value when no reasonable cast exists.
The concrete problem is to give a principled account of casting, characterizing when a cast is meaningful or safe and modeling good casting behavior with constraints, for example constraints that specify which source values must map to which target values, so that casts can be reasoned about at compile time rather than trusted blindly.

\noindent
\textbf{\emph{Data-aware compilation.}} 
The best compilation strategy may depend on the input data, target model, and prompt representation. In this paper, our compilers use fixed English-oriented renderings, but other tasks may benefit from different surface forms, including non-English or more symbolic representations. 
How compilation should adapt to data and representation remains open.

\section{Related Work}\label{sec:related}

\paragraph{Prompt Programming Languages}
Several recent works treat prompting as a \emph{language-design} problem, 
emphasizing abstraction and analyzability. 
Representative examples include $\lambda^O$~\cite{10.1145/3763143}, 
LMQL~\cite{lmql10.1145/3591300}, APPL~\cite{appldong2024}, 
and PDL~\cite{pdlvaziri2024pdldeclarativepromptprogramming}. 
$\lambda^O$ studies evaluation strategies for languages with external calls, 
including LM calls, but treats those calls as opaque effects. 
LMQL provides precise control over prompt structure and constraints, 
but programs remain closely tied to execution details. 
APPL delegates more work to the compiler, including parallelization, 
but inherits Python as its semantic substrate. 
PDL emphasizes readable specification, though with less support for rich control flow. 
We place $\DSL$ in this line of work, but with a different emphasis: 
we treat compiler design itself as semantic design, 
and use multiple compilers to make that point explicit.

\paragraph{Prompt Programming Systems}
Another line of work takes a \emph{systems} view and focuses on building reliable LM applications. 
LangChain~\cite{langchain} and LangGraph~\cite{langgraph2025} emphasize orchestration; 
DSPy~\cite{dspykhattab2024} organizes LM applications around programming, evaluation, and optimization; 
SGLang~\cite{sglang} focuses on efficient structured execution; 
and Guidance and Outlines~\cite{guidance2025,outlines2025,outlineswillard2023efficient} provide practical interfaces for constrained generation. 
Related frameworks include GenAIScript~\cite{genaiscript2025} and PromptFlow~\cite{promptflow2025}. 
These systems package effective prompting patterns into reusable abstractions, 
whereas our goal is to study what should belong in the language and what should be delegated to the compiler.

\paragraph{Constrained Decoding}
Constrained decoding is a key ingredient in many prompt systems. Existing work studies regex-based constraints~\cite{deutsch-etal-2019-general,kuchnik2023validatinglargelanguagemodels,suresh2025dingoconstrainedinferencediffusion,outlineswillard2023efficient}, CFG-based constraints~\cite{banerjee2025cranereasoningconstrainedllm,dong2025xgrammarflexibleefficientstructured,koo2024automatabased,park2025flexible,ugare2025syncode}, and more semantic search procedures such as Monte Carlo steering and backtracking~\cite{loula2025syntactic,llamppllew2023sequentialmontecarlosteering,kanda2026refinestatefficientexplorationprobabilistic,poesia2022synchromesh,ugare2025itergen}. Most of these methods constrain individual generations. By contrast, our baseline compiler derives one decidable constraint from the whole $\DSL$ program and uses it to constrain the final generation.

\paragraph{Prompt Engineering and Optimization}
Prompt engineering studies how prompt form affects model behavior, including techniques such as few-shot prompting~\cite{brown2020language,perez2021true}, chain-of-thought~\cite{wei2022chain}, self-consistency~\cite{wang2023self}, calibration~\cite{zhao2021calibrate}, and self-verification~\cite{weng2023largelanguagemodelsbetter}; see recent surveys~\cite{sahoo2025systematicsurveypromptengineering,schulhoff2024promptreportsystematicsurvey}. Prompt optimization then aims to improve prompts automatically~\cite{Ramnath_2025}, often using evaluation suites such as Holistic Evaluation~\cite{liang2022holistic}, PromptBench~\cite{zhu2023promptbench}, and ProSA~\cite{zhuo-etal-2024-prosa}. DSPy~\cite{dspykhattab2024} and \textsc{Sammo}~\cite{sammo.schnabel-neville-2024-symbolic} are especially relevant examples. We view this line of work as complementary: it asks how to find better prompts, whereas we ask what task logic should be written in the language and what execution details should be chosen by the compiler.

\section{Conclusion}\label{sec:conc}

In this work, we study prompt programming through the design and implementation of $\DSL$. Our central claim is that prompt programs should
describe tasks while remaining decoupled from lower-level execution details. We develop this view by presenting $\DSL$, a high-level
language for prompt programming, two compilers that induce different execution semantics, and a type system connected to constrained
decoding. We further implement these ideas in $\DSLPy$ and evaluate them on two case studies. Taken together, these results show that
separating prompt programs from their compilation offers a useful foundation for both the study and practice of prompt programming.

\section*{Data Availability Statement}
All data used in this paper come from publicly available benchmarks. Our statutory-reasoning experiments use SARA~\cite{DBLP:conf/kdd/HolzenbergerBD20}, 
whose data are available in the GitHub repository~\cite{sara_repo}. 
Our data labeling and transformation experiments use PromptPex~\cite{promptpexsharma2025automatictestgeneration}, 
whose data are available in the corresponding repository~\cite{promptpex_repo}. Our $\DSLPy$ implementations, evaluation scripts, and experimental results will be made available in the project repository.

\bibliographystyle{ACM-Reference-Format}
\bibliography{refs}
\newpage
\end{document}